\documentclass[twoside,11pt]{article}
\usepackage{amsmath}
\usepackage{blindtext}

\usepackage[abbrvbib, preprint]{jmlr2e}

\usepackage{lastpage}
\jmlrheading{25}{2024}{1-\pageref{LastPage}}{8/23; Revised}{}{21-0000}{Gaojie Jin, Xinping Yi, and Xiaowei Huang}

\ShortHeadings{Sample JMLR Paper}{One and Two}
\firstpageno{1}

\usepackage{hyperref}       
\usepackage{url}            
\usepackage{booktabs}       
\usepackage{amsfonts}       
\usepackage{nicefrac}       
\usepackage{microtype}      
\usepackage{xcolor}         

\usepackage{amssymb}
\usepackage{mathtools}
\usepackage{graphicx}
\usepackage[capitalize]{cleveref}
\Crefname{section}{Sec.}{Secs.}
\Crefname{table}{Tab.}{Tabs.}
\Crefname{figure}{Fig.}{Figs.}
\Crefname{thm}{Thm.}{Thms.}
\Crefname{appendix}{App.}{Apps.}

\usepackage{multirow}
\usepackage{bbm}
\usepackage{bm}
\usepackage{comment}
\usepackage{tcolorbox}
\usepackage{authblk}

\definecolor{green(pigment)}{rgb}{0.1607, 0.3843, 0.0941}
\definecolor{blue(pigment)}{rgb}{0., 0.1484, 0.6992}
\hypersetup{hidelinks}

\usepackage{caption}
\newtheorem{thm}{Theorem}[section]

\newtheorem{mypro}[thm]{Proof}
\newtheorem{mylem}[thm]{Lemma}
\newtheorem{myrem}{Remark}

\newtheorem{innercustompro}{Proof}
\newenvironment{custompro}[1]
  {\renewcommand\theinnercustompro{#1}\innercustompro}
  {\endinnercustompro}

\newtheorem{innercustomprop}{Proposition}
\newenvironment{customprop}[1]
  {\renewcommand\theinnercustomprop{#1}\innercustomprop}
  {\endinnercustomprop}

\newtheorem{innercustomthm}{Lemma}
\newenvironment{customlem}[1]
  {\renewcommand\theinnercustomthm{#1}\innercustomthm}
  {\endinnercustomthm}

\newtheorem{innercustomtheorem}{Theorem}
\newenvironment{customthm}[1]
{\renewcommand\theinnercustomtheorem{#1}\innercustomtheorem}
  {\endinnercustomtheorem}

\newcommand{\Lc}{\mathcal{L}}

\newcommand{\St}{\mathcal{S}}
\newcommand{\D}{\mathcal{D}}
\newcommand{\E}{\mathbb{E}}
\newcommand{\R}{\mathbb{R}}

\newcommand{\argmax}{\arg\max}

\newcommand{\X}{\mathcal{X}_B}
\newcommand{\Y}{\mathcal{Y}}

\newcommand{\pr}{\mathbb{P}}
\newcommand{\x}{\mathbf{x}}

\newcommand{\y}{{\mathbf{y}}}

\newcommand{\f}{\widetilde f}

\newcommand{\xx}{\boldsymbol{x}}
\newcommand{\ww}{\boldsymbol{w}}

\newcommand{\w}{\mathbf{w}}
\newcommand{\vv}{\mathbf{v}}
\newcommand{\W}{\mathbf{W}}
\newcommand{\act}{\phi}
\newcommand{\vecc}{\text{vec}}
\newcommand{\ul}{\mathbf{u}}
\newcommand{\U}{\mathbf{U}}
\newcommand{\I}{\mathbf{I}}

\newcommand{\delt}{\boldsymbol{\delta}}

\newcommand{\varp}{\bm\varepsilon}

\newcommand{\KL}{\mathrm{KL}}

\newcommand{\g}{\mathcal{G}}

\newcommand{\gaojie}[1]{{\color{red}{}#1}}
\begin{document}

\title{Reconcile Certified Robustness and Accuracy for DNN-based Smoothed Majority Vote Classifier}

\author[1*]{\textbf{Gaojie Jin}}
\author[2]{\textbf{Xinping Yi}}
\author[3]{\textbf{Xiaowei Huang}}
\affil[1]{\emph{Department of Computer Science, University of Exeter, UK}}
\affil[2]{\emph{National Mobile Communications Research Laboratory, Southeast University, China}}
\affil[3]{\emph{Department of Computer Science, University of Liverpool, UK}}
\affil[*]{\emph{Corresponding author:} 
\emph{gaojie.jin.kim@gmail.com}}

\editor{My editor}

\maketitle

\begin{abstract}
Within the PAC-Bayesian framework, the Gibbs classifier (defined on a posterior $Q$) and the corresponding $Q$-weighted majority vote classifier are commonly used to analyze the generalization performance.
However, there exists a notable lack in theoretical research exploring the certified robustness of majority vote classifier and its interplay with generalization.
In this study, we develop a generalization error bound that possesses a certified robust radius for the smoothed majority vote classifier (i.e., the $Q$-weighted majority vote classifier with smoothed inputs); In other words, the generalization bound holds under any data perturbation within the certified robust radius. 
As a byproduct, we find that the underpinnings of both the generalization bound and the certified robust radius draw, in part, upon weight spectral norm, which thereby inspires the adoption of spectral regularization in smooth training to boost certified robustness. 
Utilizing the dimension-independent property of spherical Gaussian inputs in smooth training, we propose a novel and inexpensive spectral regularizer to enhance the smoothed majority vote classifier. 
In addition to the theoretical contribution, a set of empirical results is provided to substantiate the effectiveness of our proposed method.
\end{abstract}

\begin{keywords}
  PAC-Bayes, Randomized smoothing, Generalization, Certified robustness, Spectral regularization
\end{keywords}

\section{Introduction}

Despite notable achievements have been made to improve the adversarial robustness of deep neural networks (DNNs)~\citep{papernot2016distillation,tramer2017ensemble,xu2017feature,madry2017towards,athalye2018obfuscated,wu2020skip}, it has been demonstrated that models previously deemed robust have subsequently succumbed to more powerful adversarial attacks~\citep{athalye2018obfuscated,uesato2018adversarial,croce2020reliable}.
This has motivated the need for methodologies that provide verifiable guarantees, ensuring that the predictor remains impervious to any attack within a certain perturbation radius.
Significant advancements have been achieved in the development of methodologies capable of computing certified robust radius for DNNs~\citep{katz2017reluplex,wong2018provable,wong2018scaling,huang2019achieving,jia2019certified}, but they demand comprehensive knowledge of the architecture of the predictor and pose challenges in terms of their extensibility to different models.

\begin{figure*}[t!]
\includegraphics[width=1.
\textwidth]{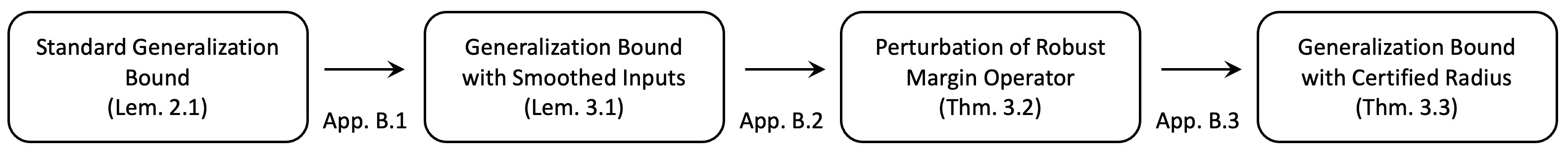}
\centering
\vspace{-5mm}
\caption{
Illustration of the theoretical framework: perturbation bound for smoothed majority vote classifiers. 
Under this framework, a standard generalization bound is extended to a smoothed generalization bound with a certified robust radius.
}
\vspace{-2mm}
\label{fig:1}
\end{figure*}

To address this problem, recent study has introduced the \emph{randomized smoothing} strategy~\citep{lecuyer2019certified,cohen2019certified,li2019certified}, an innovative approach aimed at verifying the robustness of smoothed classifiers.
Specifically, it employs the smoothing noise to the input data, followed by the determination of the most probable label by the smoothed classifier.
Then, the robust radius for the smoothed classifier can be certified.
In contrast to other methodologies, randomized smoothing stands out as an efficient and model-agnostic technique, and is highly adaptable to a wide range of machine learning models.
Although randomized smoothing provides certified guarantees for individual inputs, research on certified robustness across the entire input distribution—--particularly its interplay with generalization—--remains scarce. 
The PAC-Bayesian approach, introduced by \citet{mcallester1999pac}, is designed to provide PAC guarantees to ``Bayesian-like" learning algorithms.
Recognizing the widespread application of $Q$-weighted Majority Vote classifiers in PAC-Bayes for generalization analysis, this study makes a timely theoretical contribution. 
We introduce a margin-based generalization bound endowed with a certified robust radius for Majority Vote classifiers, thereby bridging the divide between generalization performance and robustness guarantees.
As shown in \Cref{fig:1}, the theoretical development is accomplished through a tripartite process: 
\begin{enumerate}
\item 
Building upon the properly defined expected and empirical losses of smoothed Majority Vote classifier,
we develop a margin-based PAC-Bayesian framework tailored for smoothed inputs (Lem.~\ref{lem:main1}), which bridges randomized smoothing and generalization analysis.
\item Under the margin-based framework, 
we proceed to conduct an in-depth generalization analysis of the Majority Vote classifier with a constraint on the output perturbation, relating the generalization performance to the spectral norm of weights (\Cref{thm:main2}).
\item Further to the above two steps, we derive a certified robust radius for the generalization bound of the Majority Vote classifier (\Cref{thm:main3}), which inspires us to regularize the spectral norm of weights to boost certified robustness.
\end{enumerate}

In our theoretical setting, we find the bedrock of both the generalization bound and the corresponding certified robust radius derives, in part, from the weight spectral norm. 
Therefore, we advocate the adoption of an innovative approach to regularize weight spectral norm in smooth training, where smooth training is a popular technique \citep{cohen2019certified,salman2019provably} to enhance smoothed classifiers by applying small smoothing (usually Gaussian distributed) noise to training data.
According to Gershgorin circle theorem~\citep{gershgorin1931uber}, the weight spectral norm is connected to the scale and cosine similarity of weight vectors. 
Since the scale of weights has normally been regularized by common techniques like weight decay, this work therefore centers on the regularization of weight cosine similarity.  
In smooth training with spherical Gaussian inputs, we can handily regularize cosine similarity (and weight spectral norm) by leveraging the $\ell_{1,1}$ entry-wise matrix norm of the output correlation matrix. 
This scheme offers two key advantages of effectiveness and time-efficiency. 
Through an extensive set of experiments on a wide range of datasets, we validate the usefulness of our spectral regularization method to boost the certified robustness of Majority Vote classifiers, with a little extra time consumption. 
To summarize, the contributions of this work are as follows:
\begin{itemize}
    \item By extending the concepts of margin-based generalization analysis to the smooth setting, we derive a generalization bound that possesses a certified robust radius for the Majority Vote classifier (\Cref{sec:main results}).
    To the best of our knowledge, this is the first work that synchronously studies the certified robustness and the generalization for Majority Vote classifiers.
    \item  On the strength of theoretical results, we propose an efficient method for spectral regularization to enhance the certified robustness of smoothed Majority Vote classifiers and provide comprehensive empirical results to demonstrate the effectiveness of our approach 
    (\Cref{sec:method,sec:experiments}).
\end{itemize}

\section{Preliminaries}
\label{sec:pre}
\textbf{Basic setting.}
Consider $\x \in \X$ and $y \in \Y$, where $\X=\{\x\in\mathbb{R}^d \mid \sum_{i=1}^d x_i^2\le B^2 \}$ and $\Y=\{0,1\}$.
The label $y$ will also be used as an index in the following, thus we let it be $0$ or $1$ here.
Let $\St=\{(\x_1,y_1),...,(\x_m,y_m) \}$ represent a training set consisting of $m$ samples drawn independently and identically from an underlying distribution $\mathcal{D}$, where $\mathcal{D}$ is a fixed but unknown distribution on $\X \times \Y$.
Let $f_{\w}(\x):\X\to \Y$ represent the learning function parameterized by $\w$.
Let $\mathcal{W}$ be a set of real-valued weights for functions from 
$\X$ to $\Y$, we call $\mathcal{W}$ the hypotheses of weights.
We define $f_\w$ as an $n$-layer neural network with $h$ hidden units per layer with ReLU activation function $\act(\cdot)$.
Let $\W$ and $\W_l$ denote the weight matrices of the model and the $l$-th layer, respectively.
Furthermore, let $\w$ and $\w_l$ be the vectorizations of $\W$ and $\W_l$ (i.e., $\w=\vecc(\W)$), respectively.
We can express each $f_{\w}(\x)$ as $f_{\w}(\x) = \W_{n}\act(\W_{n-1} ...\act(\W_1 \x)...)$, and define $f_\w^{(1)}(\x) = \W_{1}\x$, $f_\w^{(i)}(\x) = \W_{i}\act(f_\w^{(i-1)}(\x))$. 
For convenience, we omit the bias term as it can be incorporated into the weight matrix.
The spectral norm of matrix $\W$, denoted as $\|\W\|_2$, represents the largest singular value of $\W$, while the Frobenius norm of $\W$, denoted as $\|\W\|_F$. 
The $\ell_p$ norm of vector $\w$, denoted as $\|\w\|_p$.

 \noindent
\textbf{Margin loss.} 
The classifier $f_\w$ may exhibit different performance when evaluated on the underlying data distribution $\D$ and the training dataset $\St$.
The generalization error reflects the difference between the average losses observed in the empirical dataset and the underlying data distribution, which are estimated using the training and test datasets, respectively.
In prior works such as \cite{neyshabur2017pac,farnia2018generalizable}, the PAC-Bayesian generalization analysis for a DNN is conducted on the margin loss. 
By considering any positive margin $\gamma$, the expected margin loss is defined as 

\begin{equation}
\label{eq:marginloss}
    \Lc_{\gamma}(f_\w):=\underset{(\x,y)\sim \D}{\E}\mathbbm{1}\Big[ f_\w(\x)[y]\le \gamma+\max\limits_{j\ne y} f_\w(\x)[j] \Big],
\end{equation}
where $\mathbbm{1}[a\le b]=1$ if $a\le b$, else $\mathbbm{1}[a\le b]=0$.
To maintain consistency with prior work, we define the margin loss for both binary and multi-class settings here.
Note that setting $\gamma=0$ corresponds to the normal loss.

 \noindent
\textbf{PAC-Bayes, Gibbs classifier, and Majority Vote classifier.}
PAC-Bayes~\citep{mcallester1999pac,mcallester2003simplified} provides a tight upper bound on the generalization of a stochastic classifier known as the Gibbs classifier~\citep{laviolette2005pac}, which is defined on a posterior $Q$. 
Additionally, it furnishes a generalization guarantee for the $Q$-weighted majority vote classifier (related to this Gibbs classifier) which labels any input instance based on the most probable output of the Gibbs classifier~\citep{lacasse2006pac,germain2015risk}.
The bound is mainly determined with respect to the Kullback-Leibler divergence (KL) between the posterior $Q$ and the prior $P$ of weights (classifiers). 
We consider the posterior distribution $Q$ over classifiers of the form $f_{\w+\ul}$, where $\ul$ is a random variable whose distribution may depend on the training data.
PAC-Bayes bounds the expected loss over the posterior, i.e., the expected loss of the Gibbs classifier, 
\begin{equation}
\label{eq:pacmarginloss}
\begin{aligned}
\Lc_{\gamma}(f_{\w,\ul}):=\E_{\ul} \Big( &\underset{(\x,y)\sim \D}{\E}\mathbbm{1}\Big[ f_{\w+\ul}(\x)[y]\le \gamma+\max\limits_{j\ne y} f_{\w+\ul}(\x)[j] \Big] \Big).
\end{aligned}
\end{equation}
Given a prior distribution $P$ over classifiers that is independent of the training data, we have the following PAC-Bayesian bound.

\begin{mylem}[\cite{mcallester2003simplified,neyshabur2017pac}]
\label{lem:pac}
Consider a training dataset $\St$ with $m$ samples drawn from a distribution $\D$ with binary targets. 
Given a learning algorithm (e.g., a classifier) with prior and posterior distributions $P$ and $Q$ (i.e., $\w+\ul$) on the weights respectively, 
for any $\delta > 0$,
with probability $1-\delta$ over the draw of training data, we have that 
\begin{align}\nonumber
\Lc_{0}(f_{\w,\ul}) \le \widehat{\Lc}_{0}(f_{\w,\ul}) + 2\sqrt{\frac{2(\KL(\w+\ul || P)+\log \frac{2m}{\delta})}{m-1}},
\end{align}
where $\Lc_{0}(f_{\w,\ul})$ is the expected loss on $\D$, $\widehat{\Lc}_{0}(f_{\w,\ul})$ is the empirical loss on $\St$, and their difference yields the generalization error.
\end{mylem}


\noindent
\textbf{Smoothed Majority Vote classifier.}
In this work, we study the Majority Vote classifier, which takes a majority vote from posterior to obtain the final prediction, i.e.,
\begin{equation}
\underset{c\in\mathcal{Y}}{\arg\max} \; 
\E_{\ul} \mathbbm{1} \Big[f_{\w+\ul}(\x)[c] > \max\limits_{j\ne c} f_{\w+\ul}(\x)[j]\Big].
\end{equation}
Then, we consider the smoothed input data as $\x+\vv$, where $\vv$ is a random variable.
Following \cite{neyshabur2017pac}, we let $\ul$ be a zero mean Gaussian random variable, i.e., $\ul \sim \mathcal{N}(0,\sigma^2 \mathbf{I})$.
For convenience, we let the small perturbation $\vv$ also satisfies $\vv \sim \mathcal{N}(0,\sigma^2 \mathbf{I})$. 
\begin{myrem}
In this work, we adopt the same variance for $\ul$ and $\vv$ for both notational simplicity and experimental convenience, without compromising the generality of our theoretical findings. 
Although it is possible to choose different variances for $\ul$ and $\vv$, such variations would not affect our core theoretical results in \Cref{sec:main results}, which establishes that the weight spectral norm affects both robust generalization performance and the certified robust radius.  
\end{myrem}
Randomized smoothing constructs a smoothed classifier by enhancing a given base classifier with input noise $\mathbf{v}$,
where Gaussian distribution is a widely employed smoothing distribution in the literature~\citep{lecuyer2019certified,cohen2019certified,li2019certified}.
Typically, in these works, the smoothed classifier predicts the class with the highest confidence on the smoothed data, i.e.,
\begin{equation}
\underset{c\in\mathcal{Y}}{\arg\max} \; 
\E_{\vv} \mathbbm{1} \Big[f_{\w}(\x+\vv)[c] > \max\limits_{j\ne c} f_{\w}(\x+\vv)[j]\Big].
\end{equation}

Different from the previous work, we utilize the smoothed Majority Vote classifier $\g_{\gamma,\w}$ with the smoothing distribution on the input data and the model weights over posterior.
When queried at $\x$ over the posterior, 
the smoothed  Majority Vote classifier $\g_{\gamma,\w}$ returns the class that the base classifier is most likely to select under margin $\gamma$, i.e., 
\begin{equation}
\label{eq:smoothclassifier}
\begin{aligned}
&\g_{\gamma,\w}(\x)=\underset{c\in\mathcal{Y}}{\arg\max}
\begin{cases}
&\!\E_{\ul,\vv} \mathbbm{1} \Big[ f_{\w+\ul}(\x+\vv)[c] \!>\! \max\limits_{j\ne c} f_{\w+\ul}(\x+\vv)[j] + \gamma \Big], \quad\text{if } c=y \\
&\!\E_{\ul,\vv} \mathbbm{1} \Big[ f_{\w+\ul}(\x+\vv)[c] + \gamma \!>\! \max\limits_{j\ne c} f_{\w+\ul}(\x+\vv)[j] \Big]. \quad\text{if } c\ne y 
\end{cases}
\end{aligned}
\end{equation}
\begin{myrem}
Note that $\gamma$ is an auxiliary variable that aids in the development of the theory, which allows us to establish an upper bound on the expected loss of $\g_{0,\w}$ on $\D$.
That means we set $\gamma=0$ in the bound when the classifier processes unseen data, ensuring that it operates independently of the label.
\end{myrem}
We then define the expected loss of the smoothed  Majority Vote classifier $\g_{\gamma,\w}$ as
\begin{equation}
\label{eq:ourmarginloss}
\Lc_\gamma(\g):=\underset{(\x,y)\sim \D}{\E} \mathbbm{1} \Big[ \g_{\gamma,\w}(\x) \ne y \Big].
\end{equation}
The expected loss under $\ell_2$ norm data perturbation is defined as
\begin{equation}
\Lc_\gamma(\g,\epsilon):=\underset{(\x,y)\sim \D}{\E} \mathbbm{1} \Big[ \exists \varp \;\Big|\; \|\varp\|_2^2\le\epsilon_{\x}, \; \g_{\gamma,\w}(\x+\varp) \ne y \Big],
\end{equation}
where $\varp\in\R^d$ and $\sqrt{\epsilon_{\x}}$ is the perturbation radius for $\x$.
Here we suppose $\epsilon_{\x}$ may vary for different instances of $\x$,
which we will discuss later.
We consider $\widehat{\Lc}_{\gamma}(f_\w)$, $\widehat{\Lc}_{\gamma}(f_{\w,\ul})$, $\widehat{\Lc}_\gamma(\g)$, $\widehat{\Lc}_\gamma(\g,\epsilon)$ to be the empirical estimate of the above expected losses, e.g., $\widehat{\Lc}_\gamma(\g)$ is defined as
\begin{equation}
\widehat{\Lc}_\gamma(\g):=\frac{1}{m}\underset{(\x,y)\in \St}{\sum} \mathbbm{1} \Big[ \g_{\gamma,\w}(\x) \ne y \Big].
\end{equation}

The goal of the generalization analysis in this work is to provide theoretical comparison between the true and empirical margin losses for smoothed  Majority Vote classifiers.
Next, we show how to develop the above PAC-Bayesian generalization analysis for the smoothed  Majority Vote classifier, initially bounding $\Lc_0(\g)$ and further extending it to bound $\Lc_0(\g,\epsilon)$ with a certified robust radius.


\section{Main results}
\label{sec:main results}

\subsection{Sketch}

For clarity, we first outline our principal theoretical result (in \Cref{thm:main3}). 
Within the PAC-Bayesian framework, we formulate a generalization bound that incorporates a certified robust radius for the smoothed Majority Vote classifier. 
The primary content is elucidated as follows.
\begin{tcolorbox}
With probability at least $1-\delta$, the inequality $\Lc_0(\g,\epsilon) \leq \widehat{\Lc}_{\gamma}(\g) + \Omega_{\text{ge}}$ holds within the $\ell_2$ norm data perturbation radius $\Omega_{\text{r}}$, both $\Omega_{\text{ge}}$ and $\Omega_{\text{r}}$ are influenced, in part, by $\|\W_i\|_2$.
\end{tcolorbox}
In other words, this says both the accuracy performance and the certified robust radius, for the smoothed Majority Vote classifier, can be affected by weight spectral norm.
Consequently, we have the potential to improve certified robustness through spectral regularization. 
In the following, we provide the details of our theoretical development, as illustrated in \Cref{fig:1}.

\subsection{Elaboration}

Building upon the PAC-Bayesian framework established in Lem.~\ref{lem:pac}, which bounds the expected generalization error for Gibbs classifiers, our first objective is to develop a generalization bound specifically tailored to the smoothed Majority Vote classifier, as defined in (\ref{eq:smoothclassifier}) and (\ref{eq:ourmarginloss}). 
Drawing inspiration from the margin-based generalization analysis approach~\citep{neyshabur2017pac,bartlett2017spectrally}, we have derived a generalization bound for the smoothed Majority Vote classifier that incorporates the empirical margin loss. The details of this bound are presented below.

\begin{mylem}
\label{lem:main1}
Given Lem.~\ref{lem:pac}, 
let $f_{\mathbf{w}}: \mathcal{X}_B$ $\rightarrow \mathcal{Y}$ denote the base predictor with weights $\mathbf{w}$, and let $P$ be any prior distribution of weights that is independent of the training data, $\w+\ul$ be the posterior of weights over training dataset of size $m$.
Then, for any $\delta,\gamma>0$, and any random perturbation $\ul$, $\vv$ $s.t.$ $\pr_{\ul,\vv}(\max_{\x}|f_{\w+\ul}(\x)-f_\w(\x)|_{\infty}<\frac{\gamma}{8}\;\cap \; \max_{\x}|f_{\w+\ul}(\x+\vv)-f_\w(\x)|_{\infty}<\frac{\gamma}{8})\ge\frac{1}{2}$, 
with probability at least $1-\delta$, we have
\begin{equation}\nonumber
\Lc_{0}(\g) \leq \widehat{\Lc}_{\gamma}(\g)+4 \sqrt{\frac{\KL(\mathbf{w}+\mathbf{u} \| P)+\ln \frac{6 m}{\delta}}{m-1}},
\end{equation}
where $\Lc_{0}(\g)$ is the expected loss for the smoothed Majority Vote classifier $\g_{0,\w}$ and $\widehat{\Lc}_{\gamma}(\g)$ is the empirical estimate of the margin loss for $\g_{\gamma,\w}$.  
\end{mylem}
\textbf{Sketch of proof.} 
\emph{Lem.~\ref{lem:main1} adapts the generalization bound defined in Lem.~\ref{lem:pac}, which is originally based on normal input data, to suit a majority vote classifier over smoothed data. 
There are two main steps in the proof: the loss of the smoothed Majority Vote classifier is initially bounded by the expected margin loss of the Gibbs classifier, which is in turn bounded by the empirical margin loss of the smoothed Majority Vote classifier.
Details are given in \Cref{app:lem3.1}.}  \hfill $\square$


The methodologies employed by \citet{langford2002not,neyshabur2017exploring,dziugaite2017computing} involve utilizing PAC-Bayesian bounds to analyze the generalization behavior of neural networks, wherein the evaluation is conducted either on the KL divergence, the perturbation error $\Lc_0(f_{\w+\ul})-\Lc_0(f_{\w})$, or the entire bound numerically.
In addition, \citet{neyshabur2017pac,farnia2018generalizable} employ the PAC-Bayesian framework to derive a margin-based bound that depends on weight norms through restricting $f_{\w+\ul}(\x)-f_\w(\x)$.

In the subsequent analysis, we introduce a margin-based bound that relies on weight norms, but through imposing  restrictions on both $f_{\w+\ul}(\x)-f_\w(\x)$ and $f_{\w+\ul}(\x+\vv)-f_\w(\x)$, as demonstrated in Lem.~\ref{lem:main1}. 
This dual restriction is necessitated by the fact that the Majority Vote classifier in this work is smoothed not only over weights but also over inputs, i.e., $\x+\vv$.
Following the approach of previous studies, we choose the largest perturbation while adhering to the given constraint to obtain the conditionally tight bound. 
The following theorem provides the details of the bound, which primarily depends on the spectral norm and the Frobenius norm of weights.


\begin{thm}
\label{thm:main2}
Given Lem.~\ref{lem:main1}, for any $B, n, h > 0$, let the base classifier $f_{\mathbf{w}}: \mathcal{X}_B$ $\rightarrow \mathcal{Y}$ be an $n$-layer feedforward network with $h$ units each layer and ReLU activation function.
For any $\delta,\gamma>0$, any $\w$ over training dataset of size $m$, with probability at least $1-\delta$, we have the the following bound:
\begin{equation}\nonumber
\begin{aligned}
\Lc_{0}(\g) &\leq \widehat{\Lc}_{\gamma}(\g)+\mathcal{O}\left ( \sqrt{\frac{\Phi\left(\prod_i\|\W_i\|_2^2,\sum_i\|\W_i\|_F^2\right)+\ln \frac{nm}{\delta}}{m-1}} \right),
\end{aligned}
\end{equation}
where 
\begin{equation}\nonumber
\begin{aligned}
\Phi&\left(\prod_i\|\W_i\|_2^2,\sum_i\|\W_i\|_F^2\right) = \frac{\sum_{i}\left(\|\W_i\|_F^2 / \|\W_i\|_2^2\right)}{\Psi\left(\prod_i\|\W_i\|_2^2\right) / (\prod_i \|\W_i\|_2^2)^{\frac{1}{n}}},
\end{aligned}
\end{equation}
and
\begin{equation}\nonumber
\begin{aligned}
\Psi\left(\prod_i\|\W_i\|_2^2\right) = 
&\Bigg(\Bigg(\frac{\gamma}{2^8 n  ( h \ln (8 n h))^\frac{1}{2}\tau^{\frac{1}{2}}\prod_{i} \|\W_i\|_2^{\frac{n-1}{n}}}+\frac{B^2}{4\tau}\Bigg)^{\frac{1}{2}} - \frac{B}{2\tau^{\frac{1}{2}}}\Bigg)^2.
\end{aligned}
\end{equation} 
Here $\tau$ is the solution of $F_{\chi^2_d}(\tau)=\frac{\sqrt{2}}{2}$, where $F_{\chi^2_d}(\cdot)$ is the cumulative distribution function (CDF) for the chi-square distribution $\chi^2_d$ with $d$ degrees of freedom.
\end{thm}
\textbf{Sketch of proof.}
\emph{The main challenge \Cref{thm:main2} addressed is the computation of the KL divergence within the random perturbation limit as shown in Lem.~\ref{lem:main1}. 
Following \citet{neyshabur2017pac}, we employ a pre-determined grid method to select the prior. 
Taking into account both the sharpness limit and the Lipschitz property, the posterior can then be bounded by the weight matrices. 
These allow us to effectively bound the KL divergence, leading us to the formulation of \Cref{thm:main2}.
The detailed proof is provided in \Cref{app:lem3.2}. 
}
\hfill $\square$

The above theorem derives the generalization guarantee by restricting the variation in the output of the network and thus bounding the sharpness of the model.
This approach is similar to \citet{neyshabur2017pac,bartlett2017spectrally}, yet with a notable difference: 
the previous work depicts the change of the output with respect to random weights,
whereas we consider the variation of the output, for the Majority Vote classifier, in connection with both random weights and smoothed inputs. 

Next, we extend \Cref{thm:main2} to a certified robustness setting.  
Although randomized smoothing has been extensively explored in previous research to provide robustness guarantee for a specific input, limited attention has been given to the relationship between the certified robustness, the generalization, and the model weights.
As a first attempt, we endeavor to develop the generalization bound under a certified robust radius, i.e., 
the generalization bound holds for data perturbation within the radius.


\begin{thm}
\label{thm:main3}
Given \Cref{thm:main2}, 
for any $\x\in \X$, suppose there exist $p^A_{\w}(\x)$, $p^B_{\w}(\x)$ such that
\begin{equation}\nonumber
\begin{aligned}
    &\E_{\ul,\vv} \mathbbm{1} \Big[\underset{c}{\argmax} f_{\w+\ul}(\x+\vv)[c]=\g_{0,\w}(\x)\Big]\ge p^A_{\w}(\x) \ge p^B_{\w}(\x)\\
    &\quad\quad\quad\quad\quad\quad\quad\quad\quad\quad\quad\quad\quad\quad\quad\quad\quad\quad \ge \max_{j\ne \g_{0,\w}(\x)} \E_{\ul,\vv} \mathbbm{1} \Big[ \underset{c}{\argmax} f_{\w+\ul}(\x+\vv)[c]=j \Big].
\end{aligned}
\end{equation}
Then, for any $\delta,\gamma>0$, with probability at least $1-\delta$ we have
\begin{equation}\nonumber
\begin{aligned}
\Lc_0(\g,\epsilon) 
& \leq \widehat{\Lc}_{\gamma}(\g)+\mathcal{O}\left ( \sqrt{\frac{\Phi\left(\prod_i\|\W_i\|_2^2,\sum_i\|\W_i\|_F^2\right)+\ln \frac{nm}{\delta}}{m-1}} \right)
\end{aligned}
\end{equation}
within $\ell_2$ norm data perturbation radius $\sqrt{\epsilon_{\x}}$, where
\begin{equation}
\label{eq:radius}
\begin{aligned}
\epsilon_{\x} = &
\underbrace{-\ln\left(1-\left(\sqrt{p^A_{\w}(\x)}-\sqrt{p^B_{\w}(\x)}\right)^2\right)}_{\textbf{Model and Data Joint Dependence}} \;\cdot\;
2 \underbrace{\Psi\left(\prod_i\|\W_i\|_2^2\right)}_{\textbf{Model Dependence}}.
\end{aligned}
\end{equation}
\end{thm}
\textbf{Sketch of proof.}
\emph{ 
We draw upon the randomized smoothing framework for smoothed classifiers as established in \citet{dvijotham2020framework}, to construct a certified robust radius that complements the generalization bound in \Cref{thm:main3}. 
Details are given in \Cref{app:thm3.3}.
}
\hfill $\square$

In \Cref{thm:main3}, we utilize a commonly employed setup of randomized smoothing that there exists a lower bound $p^A_{\w}(\x)$ of the largest probability of the output and an upper bound $p^B_{\w}(\x)$ of the secondary probability of the output.
The gap between $\sqrt{p^A_{\w}(\x)}$ and $\sqrt{p^A_{\w}(\x)}$ determines the first component of the certified robust radius in (\ref{eq:radius}), which is jointly influenced by both the model and the input data.
In addition, the second component of the certified robust radius depends solely on the model, primarily on the spectral norm of weights.

The above theorem illustrates that, within our theoretical framework for the smoothed Majority Vote classifier, both the generalization bound and its corresponding certified robust radius are partially influenced by the spectral norm of weights. 
Assume other factors remain constant, reducing the spectral norm of weights can, generally and simultaneously, narrow the generalization bound and enhance the certified robust radius. 
Thus, \Cref{thm:main3} leads to our conclusion that the spectral regularization of weights may enhance the smoothed Majority Vote classifier in the aspect of certified robust radius and generalization performance. 
This theoretical analysis will guide our design of regularization in the next section for smooth training.

\begin{myrem}
These three bounds establish a framework for demonstrating how the spectral norm of weights, particularly in smoothed  Majority Vote  classifiers, impacts both generalization and the certified robust radius. 
It is well known that regularized weight spectral norm helps to reduce global Lipschitz constant of the model, and the randomized smoothing framework measures local Lipschitz constant of a model. 
Consequently, the notion that a reduced global Lipschitz constant results in improved certified robustness seems intuitive. 
Furthermore, the regularized weight spectral norm in steering the model towards a flat minimum — widely regarded as a key factor of generalization — also appears intuitive. 
Nevertheless, we would like to point out that this work develops a novel theoretical framework, mathematically encapsulating these intuitions, linking generalization and certified robustness through weight spectral norm.
To the best of our knowledge, this is the first PAC-Bayesian framework that forges a connection between generalization and certified robustness for Majority Vote classifiers, marking a significant stride in this field.
\end{myrem}


\section{Spectral regularization in smooth training}
\label{sec:method}

Previous research \citep{cohen2019certified,salman2019provably} introduces smooth training to enhance the certified robustness of smoothed classifiers through incorporating Gaussian noise into the training data. 
This work, particularly \Cref{thm:main3}, offers a different perspective: the spectral norm of the weights, as identified in the second component of (\ref{eq:radius}), may play a crucial role in both generalization and the certified robust radius.
Building upon this insight, this section explores how the spectral norm can be leveraged to augment smooth training techniques, ultimately leading to the development of a better smoothed Majority Vote classifier.


Spectral regularization has been used to improve generalization~\citep{yoshida2017spectral} and adversarial robustness~\citep{farnia2018generalizable}. 
However, given the inherent difficulty in computing spectral norm for high-dimensional matrices, it is still a costly endeavor to compute it for each layer during training.
In this work, thanks to the spherical Gaussian inputs in smooth training, we can harness the correlation matrix 
to craft an economical approach for spectral regularization as follows.

Given $l$-th weight matrix $\W_l$, according to the Gershgorin circle theorem,
we have 
\begin{equation}
\label{eq:scalecorrelation}
\begin{aligned}
\|\W_l\|_2^2 &\le \|\W_l\W_l^\top\|_{\infty}=\max_i \sum_j \Big[ \underbrace{\|\w_l^{(i)}\|_2 \|\w_l^{(j)}\|_2}_{\text{Scale}} \underbrace{|\cos(\w_l^{(i)},\w_l^{(j)})|}_{\text{Correlation}} \Big],
\end{aligned}
\end{equation}
where $\cos(\w_l^{(i)},\w_l^{(j)})=\frac{\langle\w_l^{(i)}, \w_l^{(j)}\rangle}{|\w_l^{(i)}|\cdot |\w_l^{(j)}|}$, $\w_l^{(i)}$ is the $i$-th row vector of $\W_l$.


The scale term in (\ref{eq:scalecorrelation}) has been extensively studied in previous research, particularly in seminal studies on weight decay~\citep{lecun2015deep,loshchilov2017decoupled} and batch normalization~\citep{ioffe2015batch}, among other notable works. 
In this work, given the weight scale has normally been regularized (e.g., via weight decay), we propose to leverage the correlation term in (\ref{eq:scalecorrelation}) to impose regularization on 
$\prod_i\|\W_i\|_2$.
Although minimizing the correlation term, also known as weight orthogonality~\citep{saxe2013exact,mishkin2015all,bansal2018can}, has been the subject of prior research,
our algorithm is different from previous work in two aspects: 
Firstly, the regularization is directly applied to the whole network, which is an ensemble of all layer weight matrices; 
Secondly, our algorithm is specially crafted to leverage the spherical Gaussian inputs for spectral regularization, leading to a little additional time consumption
(see~\Cref{tab:1}).

In addition, we would like to emphasize the value of optimizing the correlation term for spectral regularization. 
Suppose the weight scale in (\ref{eq:scalecorrelation}) remains constant (e.g., after weight decay), minimizing $|\cos(\w_l^{(i)},\w_l^{(j)})|$ can lead to an optimal $\|\W_l\|_2^2$. 
Specifically, for a weight matrix $\W_l$ with normalized row vectors, if $\cos(\w_l^{(i)},\w_l^{(j)})=0$ for all $i \neq j$, then $\|\W_l\|_2^2 = \|\W_l\W_l^\top\|_{\infty}$, and $\|\W_l\|_2^2$ achieves its minimum for the given scale.
Furthermore, our empirical results in \Cref{sec:Reduce spectral norm} demonstrate that regularizing the correlation term in (\ref{eq:scalecorrelation}) significantly reduces $\|\W_l\|_2$. 
In the following, we will provide the details of our novel and cheap method for spectral regularization in smooth training.


Given $\W=\W_n \W_{n-1}\cdots\W_1$, the minimal cosine similarities between row vectors of $\W$ is a straightforward sign of the optimal cosine similarities among row vectors in $\W_1,...,\W_n$.
Thus, for the reason of computational complexity, we minimize the cosine similarities of the whole weight matrix $\W$.

Let $\f_\w$ be the linear (remove non-linear components) classifier parameterized by $\W$ and $\f_\w(\x)[i]$ be the $i$-th output of $\f_\w(\x)$. 
Given $\vv\in \mathbb{R}^d$ is a spherical Gaussian, the cosine similarity between $\w^{(i)}$ and $\w^{(j)}$ (row vectors of $\W$) is equal to the statistical correlation (Pearson correlation coefficient) between  $\f_\w(\x+\vv)[i]$ and $\f_\w(\x+\vv)[j]$, i.e.,
\begin{equation}
\label{eq:cosroh}
\begin{aligned}
    &\cos(\w^{(i)},\w^{(j)}) = \rho(\f_\w(\x+\vv)[i], \f_\w(\x+\vv)[j])\\
    &=\frac{\E_{\vv}[(\f_\w(\x+\vv)[i]-\E_{\vv}(\f_\w(\x+\vv)[i]))(\f_\w(\x+\vv)[j]-\E_{\vv}(\f_\w(\x+\vv)[j]))]}
    {\sqrt{\E_{\vv}[(\f_\w(\x+\vv)[i]-\E_{\vv}(\f_\w(\x+\vv)[i]))^2]
    \E_{\vv}[(\f_\w(\x+\vv)[j]-\E_{\vv}(\f_\w(\x+\vv)[j]))^2]}}.  
\end{aligned}
\end{equation}
Note that here $\x$ is a constant vector and $\vv$ is a random vector.
Let $\mathcal{R}(\f_\w)$ be the correlation matrix of random vector $\f_\w(\x+\vv)$, i.e., $\rho(\f_\w(\x+\vv)[i], \f_\w(\x+\vv)[j])$ is the element of $i$-th row, $j$-th column of $\mathcal{R}(\f_\w)$, we have
$\sum_i\sum_j |\cos(\w^{(i)},\w^{(j)})| =   \|\mathcal{R}(\f_\w)\|_{1,1}$,
where $\|\cdot\|_{1,1}$ is the $\ell_{1,1}$ entry-wise matrix norm, representing the sum of absolute elements in the matrix. 
Therefore, to minimize the weight spectral norm through correlation term in (\ref{eq:scalecorrelation}), we can incorporate the regularization term $\|\mathcal{R}(\f_\w)\|_{1,1}$ into the objective function of smooth training, i.e., 
\begin{equation}
    \Lc(f_\w(\x+\vv),y) + \alpha \cdot \|\mathcal{R}(\f_\w)\|_{1,1},
\end{equation}
where $\alpha\in [0,+\infty)$ is a hyper-parameter to balance the relative contributions of smooth training loss $\Lc(f(\x+\vv),y)$ and regularization term $\|\mathcal{R}(\f_\w)\|_{1,1}$.
For time-efficiency, we compute $\|\mathcal{R}(\f_\w)\|_{1,1}$ once per epoch.
The next section presents comprehensive empirical evidence demonstrating that our proposed method effectively regularizes the weight spectral norm and significantly improves the performance of smoothed Majority Vote classifiers.

\section{Empirical results}
\label{sec:experiments}

Within this section, we present a comprehensive experiment to validate the effectiveness of our spectral regularization method as delineated in \Cref{sec:method}.
Our results illustrate that the spectral regularization method enhances certified robustness for smoothed Majority Vote classifiers across diverse datasets. 
Specifically, in \Cref{sec:Reduce spectral norm}, we demonstrate that our regularizer effectively reduces the weight spectral norm for Majority Vote classifiers, with a little extra time consumption. 
In \Cref{sec:exp2}, we show the reduced weight spectral norm can enhance certified robustness for smoothed Majority Vote classifiers.

\subsection{Spectral regularization}
\label{sec:Reduce spectral norm}

\begin{figure*}[t!]
\includegraphics[width=1.
\textwidth]{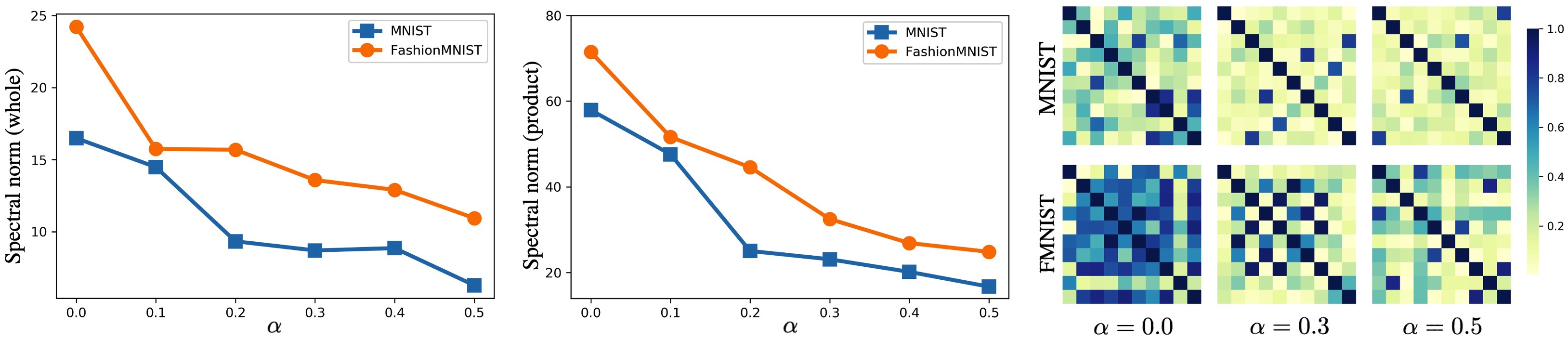}
\centering
\vspace{-7mm}
\caption{
We train the base classifier for the \emph{Majority Vote} MLPs on MNIST and FashionMNIST with $\alpha\in \{0.0, 0.1, 0.2, 0.3, 0.4, 0.5\}$, respectively. 
\textbf{Left:} the spectral norm of the whole weight matrix, i.e., $\|\W\|_2$ where $\W=\W_n \W_{n-1}\cdots\W_1$, with respect to $\alpha$.
\textbf{Middle:} the product of the spectral norms of the weight matrix, i.e., $\prod_i \|\W_i\|_2$, with respect to $\alpha$.
\textbf{Right:} the cosine similarity matrix of row vectors of the whole weight matrix $\W$.
}
\vspace{0mm}
\label{fig:2}
\end{figure*}

Since the weight spectral norm of small multilayer perceptron (MLP) can be easily computed, we implement MLPs with different values of hyper-parameter $\alpha$ on MNIST and FashionMNIST to show the efficacy of our approach in reducing weight spectral norm.
As shown in \Cref{fig:2}, our method can effectively reduce the whole weight matrix spectral norm ($\|\W\|_2$) and the product of weight spectral norms ($\prod_i \|\W_i\|_2$).
Both of them significantly decrease as $\alpha$ increases.
However, large $\alpha$ can also damage the training accuracy of the classifier, thus we set $\alpha=0.1,\; 0.2$ as the default value for the following experiments in \Cref{sec:exp2}.
Moreover, our method incurs only a minor overhead in training time due to the ease of computing the correlation matrix in a linear network, as demonstrated in \Cref{tab:1}.
We provide the details of the models in \Cref{app:model}.

While this paper does not focus on effective and efficient methods for estimating the spectral norm of neural networks, it is worth noting that several studies, including \citet{sedghi2018singular,bibi2019deep,yi2022asymptotic}, have proposed various approaches to address this challenge. 
Our primary emphasis is on elucidating the relationship between generalization and certified robust radius for Majority Vote classifiers, as illustrated in \Cref{fig:1}.


\begin{table*}[t!]
\centering
\caption{Running time (seconds/epoch) for MLP on MNIST, ResNet110 on CIFAR-10 and ResNet50 on ImageNet.}
\label{tab:1}
\vspace{-1mm}
\scalebox{0.95}{
\begin{tabular}{ccccccccccc}
\specialrule{.1em}{.075em}{.075em} 
\multicolumn{3}{c}{MLP (MNIST)} && \multicolumn{3}{c}{ResNet110 (CIFAR-10)} && \multicolumn{3}{c}{ResNet50 (ImageNet)}        
\\
Normal & Spectral & Gap && Normal & Spectral & Gap && Normal & Spectral & Gap         
\\
\cline{1-3} \cline{5-7} \cline{9-11} 
17.0 & 17.1 & +0.5\% && 32.6 & 33.7 & +3.3\% && 8970 & 9160 & +2.1\%     \\
\specialrule{.1em}{.075em}{.075em}
\end{tabular}
}
\vspace{0mm}
\end{table*}

\subsection{Enhancing certified robustness}
\label{sec:exp2}

\textbf{Certified test accuracy.} In adversarially robust classification, one important metric is the certified test accuracy at radius $r_{test}$. 
For smoothed Majority Vote classifiers, this metric is defined as the fraction of the test set that $\g_{0,\w}$ classifies correctly with a prediction that is certifiably robust within an $\ell_2$ norm ball of radius $r_{test}$. 
Let $\mathcal{S}_ {test}$ be the test set, $m_ {test}$ be the size of test set, and $r_{test}$ be the $\ell_2$ norm perturbation radius, the certified test error (1 - accuracy) under $\ell_2$ norm perturbation radius $r_{test}$ is defined as $$\frac{1}{m_ {test}}\underset{({\bf x},y) \in \mathcal{S}_ {test}}{\sum} \mathbbm{1} \Big[ \exists \varp \Big| \|\varp\|_2\le r_{test}, \mathcal{G} _ {0, {\bf w}}({\bf x}+\varp) \ne y \Big].$$

\noindent\textbf{Variance for smothed Majority Vote classifiers.}
We utilize a sharpeness-like method \citep{keskar2016large,jiang2019fantastic} to get the largest $\sigma$ and adopt it in classifiers, which can be tolerant of about $2\%$ generalization drop for MNIST and about $5\%$ generalization drop for other datasets.
Details of the method are given in \Cref{app:sharpness}.

\noindent\textbf{Certified radius.} 
It is difficult to directly compute the above certified test accuracy.
Thus, in this work, our primary evaluation algorithm closely adheres to \citet{cohen2019certified}, to estimate certified radius $R$ for each test sample, it allows us to evaluate certified test error (accuracy) under any radius.
Our evaluation method differs from previous randomized smoothing work in that we use a certified framework that is tailored to the Majority Vote classifier. 
Specifically, given an input $\x$, the certified robust radius for the Majority Vote classifier can be estimated through
$R = (-2\sigma^2 \ln(1-(\sqrt{p^A_{\w}(\x)}-\sqrt{p^B_{\w}(\x)})^2))^{\frac{1}{2}}$,
where $p^A_{\w}(\x)$, $p^B_{\w}(\x)$ share the same definitions as outlined in \Cref{thm:main3}, and $\sigma^2$ represents the variance of the perturbation.
We provide the details in \Cref{app:radius}.

\noindent
\textbf{Running setting.}
When running the evaluation algorithm to compute $p^A_{\w}(\x)$ and $p^B_{\w}(\x)$, we use $100$ Monte Carlo samples for selection, $100,000$ samples for estimation on MNIST and FashionMNIST, $10,000$ samples for estimation on CIFAR-10~\citep{krizhevsky2009learning} and ImageNet~\citep{deng2009imagenet}.

Note that our experiments are not comparable with those in \citet{cohen2019certified,salman2019provably,zhai2019macer}, as their evaluations are tailored for standard smoothed classifiers, whereas ours are specifically designed for smoothed $Q$-weighted Majority Vote classifiers.

\begin{figure*}[t!]
\includegraphics[width=1.
\textwidth]{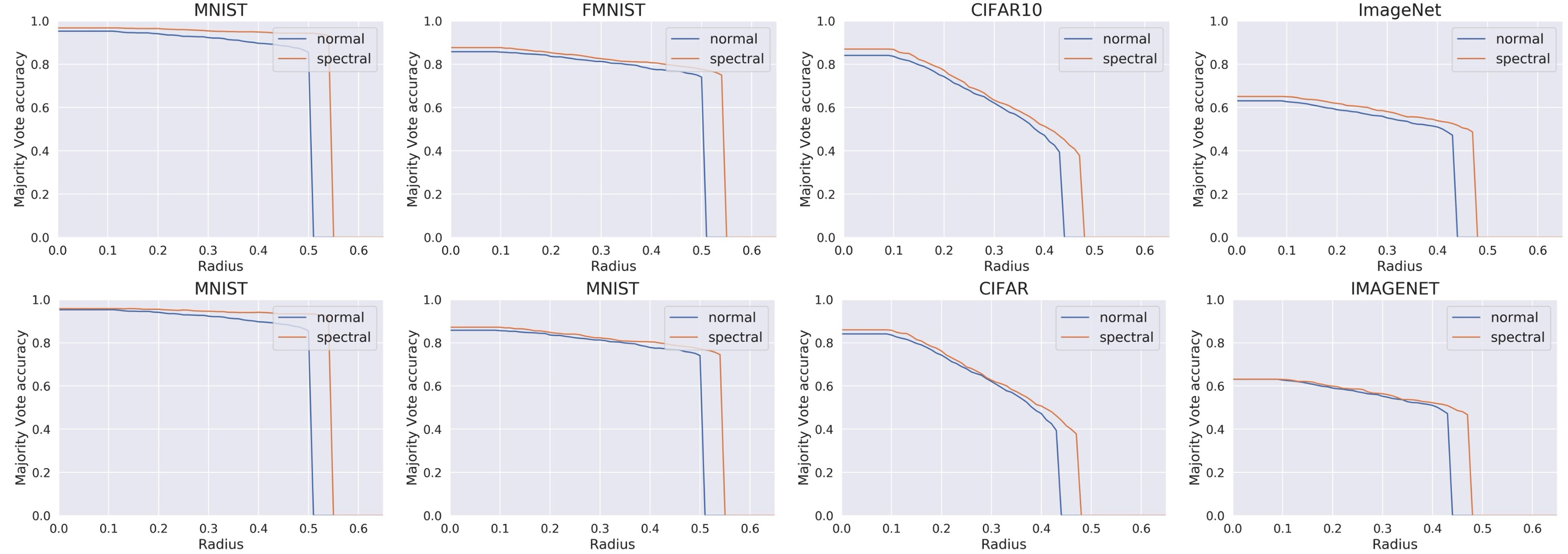}
\centering
\vspace{-7mm}
\caption{
Experiments on the \emph{Majority Vote classifiers}, for MLP on MNIST, MLP on FashionMNIST, ResNet110 on CIFAR-10, and ResNet50 on ImageNet, respectively.
We certify the full MNIST, FashionMNIST, CIFAR-10 test sets and a subsample of 1000 examples from the ImageNet test set.
Top: $\alpha=0.1$; Bottom: $\alpha=0.2$.
}
\vspace{-4mm}
\label{fig:3}
\end{figure*}


\Cref{fig:3} illustrates the upper envelope of the smoothed Majority Vote classifier accuracies for MLPs trained on MNIST and FashionMNIST, as well as ResNet110 on CIFAR-10 and ResNet50 on ImageNet. 
As shown in the figure, the application of spectral regularization consistently enhances certified robustness across all four datasets. 
The Majority Vote classifiers, when subjected to spectral regularization, not only maintain high accuracy but also achieve a modest increase in the certified robust radius.

\vspace{-2mm}
\section{Conclusion}
\vspace{-1mm}

In this paper, we explored the theoretical connection between the generalization performance, the certified robust radius, and the model weights for smoothed Majority Vote classifiers, leading to the development of a margin-based generalization bound within our smooth framework. 
As a result of these theoretical findings, we were inspired to formulate a highly effective and efficient spectral regularizer tailored for smoothed Majority Vote classifiers. 
Subsequently, we conducted extensive experiments to empirically demonstrate the effectiveness of our spectral regularizer in normalizing weight spectral norms and improving certified robustness.

\appendix

\section{Discussion of related work}
\label{app:morerelated}
\textbf{PAC-Bayes} is a general framework that enables efficient study of generalization for various machine learning models.
Numerous canonical works have been developed in the past decades since the inception of PAC-Bayes by \citet{mcallester1999pac}.
E.g., \citet{seeger2002pac} studies PAC-Bayes for Gaussian process classification; 
\citet{langford2002not} uses PAC-Bayes to bound the true error rate of a continuous valued classifier; 
\citet{langford2003pac} develops a margin-based PAC-Bayesian bound which is more data-dependent;
\citet{germain2009pac} proposes different learning algorithms to find linear classifiers that minimize PAC-Bayesian bounds;
\citet{parrado2012pac} explores the capabilities of PAC-Bayes to provide tight predictions for the generalization of SVM;
\citet{alquier2016properties} shows that when the risk function is convex, a variational approximation of PAC-Bayesian posterior can be obtained in polynomial time;
\citet{thiemann2017strongly} proposes a PAC-Bayesian bound and a way to construct a hypothesis space, so that the bound is convex in the posterior distribution;
\citet{dziugaite2018data} shows a differentially private data-dependent prior can yield a valid PAC-Bayesian bound;
\citet{letarte2019dichotomize} studies PAC-Bayes for multilayer neural networks with binary activation; 
\citet{rivasplata2020pac} provides a basic PAC-Bayesian inequality for stochastic kernels;
\citet{Gaojie2020,jin2025enhancing,11027475} studies PAC-Bayes with non-spherical Gaussian posterior distribution for DNNs;
\citet{perez2021tighter} conducts an empirical study on the use of training objectives derived from PAC-Bayesian bounds to train probabilistic neural networks;
\citet{dziugaite2021role} shows that in some cases, a stronger PAC-Bayesian bound can be obtained by using a data-dependent oracle prior;
\citet{lotfi2022pac} develops a compression approach to provide tight PAC-Bayesian generalization bounds on a variety of tasks, including transfer learning; 
\citet{haddouche2022online} proves new PAC-Bayesian bounds in the online learning framework;
\citet{amit2022integral} presents a PAC-Bayesian generalization bound which enables the replacement of the KL divergence with a variety of Integral Probability Metrics;
\citet{wu2022split} derives a PAC-Bayes-split-kl inequality to bound the expected loss;
\citet{biggs2023tighter} introduces a modified version of the excess risk to obtain empirically tighter, faster-rate PAC-Bayesian generalization bounds.
\citet{livni2020limitation} presents a limitation for the PAC-Bayes and demonstrates an easy learning task that is not amenable to a PAC-Bayesian analysis.

\citet{viallard2021pac} introduces a general PAC-Bayesian generalization framework for adversarial robustness, providing an estimate of model's invariance in majority votes to imperceptible input perturbations at test time. 
The approach consists in considering an averaged adversarial robustness risk, which represents the probability of the model incorrectly classifying a perturbed example (this can be viewed as an average risk across perturbations). To be more informative, \citet{viallard2021pac} further proposes an averaged-max adversarial risk. 
This metric evaluates the probability of at least one perturbation leading to misclassification. Then, for each adversarial risk, \citet{viallard2021pac} derives a PAC-Bayesian framework to bound the averaged risk on the perturbations for majority votes (over the whole class of hypotheses). 
These bounds are applicable to any form of attack and can be directly optimized to obtain a robust model.


In contrast to previous research, this paper investigates the theoretical properties of smoothed Majority Vote classifiers within the PAC-Bayesian framework.
As shown in \Cref{fig:1}, the most notable difference is that this work introduces a margin-based generalization bound tailored for the smoothed Majority Vote classifier with a certified robust radius.

Following the groundbreaking research of \textbf{randomized smoothing} by~\citet{lecuyer2019certified,cohen2019certified,li2019certified}, several notable contributions have emerged.
\citet{kumar2020curse} shows that extending the smoothing technique to defend against other attack models can be challenging.
\citet{lee2019tight,schuchardt2020collective,wang2021certified} introduce extensions of randomized smoothing that address discrete perturbations like  $\ell_0$-perturbations, whereas \citet{bojchevski2020efficient,gao2020certified,levine2020randomized,liu2021pointguard} delve into extensions related to graphs, patches, and point cloud manipulations.
\citet{yang2020randomized,dinghuaizhang} derive methods to determine certificates with $\ell_1, \ell_2$, and $\ell_\infty$ norms.
\citet{xie2021crfl} trains certifiably robust federated learning models against backdoor attacks via randomized smoothing.
\citet{mu2023certified} leverages randomized smoothing for multi-agent reinforcement learning system to determine actions with guaranteed certified bounds.
\citet{dvijotham2020framework} offers theoretical derivations applicable to both continuous and discrete smoothing measures, whereas \citet{mohapatra2020higher} enhances certificates by incorporating gradient information and \citet{horvath2021boosting} uses ensembles to improve certificates.
\citet{rosenfeld2020certified,levine2020deep,jia2021intrinsic,weber2023rab} use randomized smoothing to defend against data poisoning attacks.
In addition to norm-balls certificates, \citet{fischer2020certified,li2021tss} demonstrate how randomized smoothing can be used to certify geometric operations like rotation and translation.
In their respective works, \citet{chiang2020detection,fischer2021scalable} show how certificates can be extended from classification to regression (and object detection) and segmentation.
In the context of classification, \citet{jia2019certified} broadens the scope of certificates by encompassing not only the top-1 class but also the top-k classes, whereas \citet{kumar2020certifying} focuses on certifying the confidence of the classifier, extending beyond solely the top-class prediction.
\citet{hong2022unicr} proposes a universally approximated certified robustness framework to provide robustness certification for any input data on any classifier;
\citet{hao2022gsmooth} proposes generalized randomized smoothing to certify robustness against general semantic transformations.
\citet{schuchardt2022invariance} proposes a gray-box approach to enhance randomized smoothing technique with white-box knowledge about invariances.
\citet{mehra2021robust} presents a novel bilevel optimization-based poisoning attack that damages the robustness of certified robust models.
\citet{alfarra2022data} shows that the variance of the Gaussian distribution in randomized smoothing can be optimized at each input so as to maximize the certification radius.
\citet{mohapatra2021hidden} shows that the decision boundaries of smoothed classifiers will shrink, leading to disparity in class-wise accuracy.
\citet{alfarra2022deformrs} reformulates the randomized smoothing framework which can scale to large networks on large input datasets.
\citet{chen2022input} proposes Input-Specific Sampling acceleration to achieve the cost-effectiveness for randomized smoothing robustness certification.

In contrast to previous work on randomized smoothing, this paper explores the theoretical properties of smoothed Majority Vote classifiers under PAC-Bayesian framework.
Specifically, it offers a theoretical explanation of how a smoothed Majority Vote classifier can achieve superior generalization performance and a larger certified robust radius.

\section{Proofs}
\label{app:prooff}

\subsection{Proof for Lem.~\ref{lem:main1}}
\label{app:lem3.1}
First, we restate Lem.~\ref{lem:main1} here.

\begin{customlem}{3.1}
Given Lem.~\ref{lem:pac}, 
let $f_{\mathbf{w}}: \mathcal{X}_B$ $\rightarrow \mathcal{Y}$ denote the base predictor with weights $\mathbf{w}$ over training dataset of size $m$, and let $P$ be any prior distribution of weights that is independent of the training data, $\w+\ul$ be the posterior of weights.
Then, for any $\delta,\gamma>0$, and any random perturbation $\ul$, $\vv$ $s.t.$ $\pr_{\ul,\vv}(\max_{\x}|f_{\w+\ul}(\x)-f_\w(\x)|_{\infty}<\frac{\gamma}{8}\;\cap \; \max_{\x}|f_{\w+\ul}(\x+\vv)-f_\w(\x)|_{\infty}<\frac{\gamma}{8})\ge\frac{1}{2}$, 
with probability at least $1-\delta$, we have
\begin{equation}\nonumber
\Lc_{0}(\g) \leq \widehat{\Lc}_{\gamma}(\g)+4 \sqrt{\frac{\KL(\mathbf{w}+\mathbf{u} \| P)+\ln \frac{6 m}{\delta}}{m-1}},
\end{equation}
where $\Lc_{0}(\g)$ is the expected loss for the smoothed Majority Vote classifier $\g_{0,\w}$ and $\widehat{\Lc}_{\gamma}(\g)$ is the empirical estimate of the margin loss for $\g_{\gamma,\w}$.  
\end{customlem}

In the following, we provide the details of the proof.

\begin{custompro}{for Lem.~\ref{lem:main1}}
Let $\St_{\ul,\vv}$ be the set of perturbations with the following property:
\begin{equation}
\label{eq:su}
\mathcal{S}_{\mathbf{u},\vv} \subseteq\left\{(\mathbf{u},\vv)\Big|\max _{\mathbf{x} \in \mathcal{X}_B}| f_{\mathbf{w+u}}(\mathbf{x+v})-\left.f_{\mathbf{w}}(\mathbf{x})\right|_{\infty}<\frac{\gamma}{8} \;\cap\; 
\max _{\mathbf{x} \in \mathcal{X}_B}| f_{\mathbf{w+u}}(\mathbf{x})-\left.f_{\mathbf{w}}(\mathbf{x})\right|_{\infty}<\frac{\gamma}{8}
\right\}.
\end{equation}
Let $q$ be the joint probability density function over $\ul, \vv$. 
We construct a new distribution $\tilde Q$ over $\tilde \ul, \tilde \vv$ that is restricted to $\St_{\ul,\vv}$ with the probability density function:
\begin{equation}
\label{eq:qu}
\tilde{q}(\tilde{\mathbf{u}},\tilde{\mathbf{v}})= \begin{cases} \frac{1}{z} q(\tilde{\mathbf{u}},\tilde{\mathbf{v}}) & \tilde{\mathbf{u}},\tilde{\mathbf{v}} \in \mathcal{S}_{\mathbf{u},\vv}, \\ 0 & \text { otherwise}, \end{cases}
\end{equation}
where $z$ is a normalizing constant and by the lemma assumption $z=\pr((\ul,\vv)\in \St_{\ul,\vv})\ge\frac{1}{2}$.
By the definition of $\tilde Q$, we have:
\begin{equation}
\max _{\mathbf{x} \in \mathcal{X}_B} | f_{\mathbf{w}+\tilde \ul}(\mathbf{x}+\tilde \vv)-\left.f_{\mathbf{w}}(\mathbf{x})\right|_{\infty}<\frac{\gamma}{8} \;\text{and}\;
\max _{\mathbf{x} \in \mathcal{X}_B} | f_{\mathbf{w}+\tilde \ul}(\mathbf{x})-\left.f_{\mathbf{w}}(\mathbf{x})\right|_{\infty}<\frac{\gamma}{8}.
\end{equation}
Therefore, with probability at least $1-\delta$ over training dataset $\St$, we have:
\begin{equation}\nonumber
\begin{aligned}
\Lc_0(\g) &\le \Lc_{\frac{\gamma}{2}}(f_{\w,\tilde \ul}) \quad\quad\quad\quad\quad\quad & \triangleright \text{because of Pf.~\ref{lem:app1}} & \\
&\le \widehat{\Lc}_{\frac{\gamma}{2}}(f_{\w,\tilde\ul}) + 2\sqrt{\frac{2(\KL(\w+\tilde\ul || P)+\log \frac{2m}{\delta})}{m-1}}   & \triangleright \text{because of Lem.~\ref{lem:pac}} & \\
&\le \widehat{\Lc}_{\gamma}(\g) + 2\sqrt{\frac{2(\KL(\w+\tilde\ul || P)+\log \frac{2m}{\delta})}{m-1}}   & \triangleright \text{because of Pf.~\ref{lem:app2}} & \\
&\le \widehat{\Lc}_{\gamma}(\g) + 4\sqrt{\frac{\KL(\w+\ul || P)+\log \frac{6m}{\delta}}{m-1}}  & \triangleright \text{because of Pf.~\ref{lem:app3}} & 
\end{aligned}
\end{equation}
Hence, proved. \hfill $\square$
\end{custompro}

\begin{mypro}
\label{lem:app1}
Given (\ref{eq:su}) and (\ref{eq:qu}), for all $(\tilde \ul, \tilde\vv) \in \tilde Q$, we have
\begin{equation}
\label{eq:app1.1}
\begin{aligned}
&\max_{\x\in\X}|f_{\mathbf{w}+\tilde\ul}(\mathbf{x}+\tilde \vv)-f_{\mathbf{w}}(\mathbf{x})|_\infty<\frac{\gamma}{8},\\
&\max_{\x\in\X}|f_{\mathbf{w}+\tilde\ul}(\mathbf{x})-f_{\mathbf{w}}(\mathbf{x})|_\infty<\frac{\gamma}{8}.
\end{aligned}
\end{equation}
For all $\x\in \X$ s.t. $\g_{0,\w}(\x)\ne y$, as $z=\pr((\ul,\vv)\in \St_{\ul,\vv})\ge\frac{1}{2}$, there exists $(\check\ul,\check\vv) \in \tilde Q$ s.t.
\begin{equation}
f_{\mathbf{w}+\check\ul}(\mathbf{x}+\check \vv)[\g_{0,\w}(\x)]>f_{\mathbf{w}+\check \ul}(\mathbf{x}+\check \vv)[y].    
\end{equation}

Then for all $(\tilde \ul, \tilde \vv) \in \tilde Q$,  all $\x\in \X$ s.t. $\g_{0,\w}(\x)\ne y$, we have 
\begin{equation}
\begin{aligned}
f_{\mathbf{w}+\tilde \ul}(\mathbf{x})[\g_{0,\w}(\x)]+\frac{\gamma}{4} &> f_{\mathbf{w}}(\mathbf{x})[\g_{0,\w}(\x)]+\frac{\gamma}{8}\\
&> f_{\mathbf{w}+\check\ul}(\mathbf{x}+\check \vv)[\g_{0,\w}(\x)]\\
&>f_{\mathbf{w}+\check \ul}(\mathbf{x}+\check \vv)[y]\\
&>f_{\mathbf{w}}(\mathbf{x})[y]-\frac{\gamma}{8}\\
&>f_{\mathbf{w}+\tilde \ul}(\mathbf{x})[y]-\frac{\gamma}{4}.
\end{aligned}
\end{equation}
Thus we have $\Lc_0(\g) \le \Lc_{\frac{\gamma}{2}}(f_{\w,\tilde \ul})$. 
\hfill $\square$
\end{mypro}

\begin{mypro}
\label{lem:app2}
Given  (\ref{eq:su}), (\ref{eq:qu}) and (\ref{eq:app1.1}), for all $(\tilde \ul, \tilde \vv) \in \tilde Q$, 
if there exists $\x\in\X$ and $(\check \ul, \check \vv) \in \tilde Q$ s.t. 
$f_{\w+\check \ul}(\x)[y]<\max_{j\ne y}f_{\w+\check \ul}(\x)[j]+\frac{\gamma}{2}$, 
we have
\begin{equation}
\begin{aligned}
f_{\mathbf{w}+\tilde \ul}(\mathbf{x}+\tilde \vv)[j]+\frac{3\gamma}{4}&>f_\w(\x)[j]+\frac{5\gamma}{8}\\
&>f_{\w+\check \ul}(\x)[j]+\frac{\gamma}{2}\\
&>f_{\w+\check \ul}(\x)[y]\\
&>f_{\w}(\x)[y]-\frac{\gamma}{8}\\
&>f_{\mathbf{w}+\tilde \ul}(\mathbf{x}+\tilde \vv)[y]-\frac{\gamma}{4}.
\end{aligned}
\end{equation}
Hence, if there exists $\x\in\X$ and $(\check \ul, \check \vv) \in \tilde Q$ s.t. 
$f_{\w+\check \ul}(\x)[y]<\max_{j\ne y}f_{\w+\check \ul}(\x)[j]+\frac{\gamma}{2}$, we have 
$f_{\mathbf{w}+\tilde \ul}(\mathbf{x}+\tilde \vv)[j]+\gamma>f_{\mathbf{w}+\tilde \ul}(\mathbf{x}+\tilde \vv)[y]$ for all $(\tilde \ul, \tilde \vv) \in \tilde Q$. 
Given $z\ge \frac{1}{2}$, we have $\g_{\gamma,\w}(\x)\ne y$. 
Thus, we have $\widehat\Lc_{\gamma}(\g)\ge \widehat\Lc_{\frac{\gamma}{2}}(f_{\w,\tilde \ul})$.
\hfill $\square$
\end{mypro}

\begin{mypro}
\label{lem:app3}
Given $q$, $\tilde q$, $z$ and $\St_{\ul,\vv}$ in (\ref{eq:qu}),
let $\St_{\ul,\vv}^c$ denote the complement set of $\St_{\ul,\vv}$ and $\tilde q^c$ denote the normalized density function restricted to $\St_{\ul,\vv}^c$. 
Then, we have
\begin{equation}
\KL(q\| p) = z\KL(\tilde q\| p) + (1-z)\KL(\tilde q^c\| p)-H(z),
\end{equation}
where $H(z)=-z \ln z-(1-z) \ln (1-z) \leq 1$ is the binary entropy function. 
Since $\KL$ is always positive, we get 
\begin{equation}
\KL(\tilde{q} \| p)=\frac{1}{z}[\KL(q \| p)+H(z))-(1-z) \KL(\tilde{q}^c \| p)] \leq 2(\KL(q \| p)+1).
\end{equation}
Thus we have $2(\KL(\w+\ul || P)+\log \frac{6m}{\delta})\ge \KL(\w+\tilde\ul || P)+\log \frac{2m}{\delta}$.
\hfill $\square$
\end{mypro}

\hspace*{\fill}

\hspace*{\fill}

\subsection{Proof for \Cref{thm:main2}}
\label{app:lem3.2}
First, we restate \Cref{thm:main2} here.
\begin{customthm}{3.2}
Given Lem.~\ref{lem:main1}, for any $B, n, h > 0$, let the base classifier $f_{\mathbf{w}}: \mathcal{X}_B$ $\rightarrow \mathcal{Y}$ be an $n$-layer feedforward network with $h$ units each layer and ReLU activation function.
Choose the largest perturbation under the restriction, for any $\delta,\gamma>0$, any $\w$ over training dataset of size $m$, with probability at least $1-\delta$, we have the the following bound:
\begin{equation}\nonumber
\Lc_{0}(\g) \leq \widehat{\Lc}_{\gamma}(\g)+\mathcal{O}\left ( \sqrt{\frac{\Phi\left(\prod_i\|\W_i\|_2^2,\sum_i\|\W_i\|_F^2\right)+\ln \frac{nm}{\delta}}{m-1}} \right),
\end{equation}
where 
\begin{equation}\nonumber
\Phi\left(\prod_i\|\W_i\|_2^2,\sum_i\|\W_i\|_F^2\right) = \frac{\sum_{i}\left(\|\W_i\|_F^2 / \|\W_i\|_2^2\right)}{\Psi\left(\prod_i\|\W_i\|_2^2\right) / (\prod_i \|\W\|_2^2)^{\frac{1}{n}}},
\end{equation}
and
\begin{equation}\nonumber
\Psi\left(\prod_i\|\W_i\|_2^2\right) = 
\left(\left(\gamma\Bigg/\left(2^8 n  ( h \ln (8 n h))^\frac{1}{2}\tau^{\frac{1}{2}}\prod_{i} \|\W_i\|_2^{\frac{n-1}{n}}\right)+\frac{B^2}{4\tau}\right)^{\frac{1}{2}} - \frac{B}{2\tau^{\frac{1}{2}}}\right)^2.
\end{equation} 
Here $\tau$ is the solution of $F_{\chi^2_d}(\tau)=\frac{\sqrt{2}}{2}$, where $F_{\chi^2_d}(\cdot)$ is the CDF for the chi-square distribution $\chi^2_d$ with $d$ degrees of freedom.
\end{customthm}
In the following, we provide the details of the proof.

\begin{custompro}{for \Cref{thm:main2}.}
Following \citet{neyshabur2017pac}, we use two main steps to prove \Cref{thm:main2}. 
Firstly, utilizing Lems.~\ref{lem:peturbound} and \ref{lam:perbound2}, we compute the maximum allowable perturbation of $\ul$ required to satisfy the given condition on the margin $\gamma$.  
In the second step, we compute the KL term in Lem.~\ref{lem:main1}, considering the perturbation obtained from the previous step. 
This computation is essential in deriving the PAC-Bayesian bound.

Consider a network with weights $\W$ that we can regularize through dividing each weight matrix $\W_i$ by its spectral norm $\|\W_i\|_2$. 
Let $\beta$ be defined as the geometric mean of the spectral norms of all weight matrices, i.e., $\beta=\left(\prod_{i=1}^n\left\|\W_i\right\|_2\right)^{1 / n}$. 
We introduce a modified version of the weights, denoted as $\widetilde{\W}_i=\frac{\beta}{\left\|\W_i\right\|_2} \W_i$, which is obtained by scaling the original weights $\W_i$ by a factor of $\frac{\beta}{\left\|\W_i\right\|_2}$. 
As a consequence of the homogeneity property of ReLU, the behavior of the network with the modified weights $f_{\tilde \w}$, is the same as the original network $f_\w$.

Moreover, we find that the product of the spectral norms of the original weights, $\left(\prod_{i=1}^n\left\|\W_i\right\|_2\right)$, is equal to the product of the spectral norms of the modified weights, $\left(\prod_{i=1}^n\left\|\widetilde{\W}_i\right\|_2\right)$. 
Additionally, the ratio of the Frobenius norm of the original weights to their spectral norm is equal to the ratio of the modified weights, i.e., $\frac{\left\|\W_i\right\|_F}{\left\|\W_i\right\|_2}=\frac{\left\|\tilde{\W}_i\right\|_F}{\left\|\tilde{\W}_i\right\|_2}$. 
Consequently, the excess error mentioned in the Theorem statement remains unchanged under this weight normalization. 
Hence, it suffices to prove the Theorem only for the normalized weights $\tilde \w$. 
We assume, without loss of generality, that the spectral norm of each weight matrix is equal to $\beta$, i.e., for any layer $i$, $\|\W_i\|_2 = \beta$.


In our approach, we initially set the prior distribution $P$ as a Gaussian distribution with zero mean and diagonal covariance matrix $\sigma^2 \mathbf{I}$.
We incorporate random perturbations $\ul, \vv\sim\mathcal{N}(0,\sigma^2 \mathbf{I})$, with $\sigma$'s value to be determined in relation to $\beta$ at a later stage. 
Specifically, since the prior must be independent of the learned predictor $\w$ and its norm, we choose $\sigma$ of prior according to an estimated value $\tilde \beta$. 
We calculate the PAC-Bayesian bound for each $\tilde{\beta}$ selected from a pre-determined grid, offering a generalization guarantee for all $\mathbf{w}$ satisfying $|\beta-\tilde{\beta}| \leq \frac{1}{n} \beta$. 
This ensures coverage of each relevant $\beta$ value by some grid's $\tilde{\beta}$. 
Subsequently, we apply a union bound across all grid-defined $\tilde{\beta}$. 
For now, we will consider a set $\tilde{\beta}$ and the corresponding $\w$ that meet the condition $|\beta-\tilde{\beta}| \leq \frac{1}{n} \beta$, implying $\frac{1}{e} \beta^{n-1} \leq \tilde{\beta}^{n-1} \leq e \beta^{n-1}$.

According to \cite{bandeira2021spectral} and $\mathbf{u} \sim \mathcal{N}\left(0, \sigma^2 \mathbf{I}\right)$, we can get the following bound for $\|\U_i\|_2$:
\begin{equation}
\label{eq:uwbound}
\mathbb{P}_{\ul_i \sim \mathcal{N}\left(0, \sigma^2 \I\right)}\left[\left\|\U_i\right\|_2>t\right] \leq 2 h e^{-t^2 / 2 h \sigma^2} \text {. }
\end{equation}
By taking a union bound over the layers, we can establish that, with a probability $\geq \frac{\sqrt{2}}{2}$, the spectral norm of the perturbation $\U_i$ in each layer is bounded by $\sigma \sqrt{2 h \ln ((4+2\sqrt{2}) n h)}$.

As $\vv\sim \mathcal{N}(0,\sigma^2\mathbf{I})$ and takes value in $\R^d$, $\frac{\|\vv\|_2^2}{\sigma^2}$ has a chi-square distribution $\chi^2_d$.
Let $F_{\chi^2_d}(\cdot)$ be the cumulative distribution function of $\chi^2_d$, and $F_{\chi^2_d}(\tau)=\frac{\sqrt{2}}{2}$.
Then, with a probability $\frac{\sqrt{2}}{2}$, $\|\vv\|_2$ is upper bounded by $\sqrt{\tau}\sigma$.

Thus, with probability at least $\frac{1}{2}$ (i.e., $\frac{\sqrt{2}}{2}\cdot \frac{\sqrt{2}}{2}$), the above bounds can both hold.
Plugging the bounds into Lems.~\ref{lem:peturbound} and \ref{lam:perbound2}, we have that 
\begin{equation}
\label{eq:cod1}
\begin{aligned}
\max _{\mathbf{x} \in \mathcal{X}_B}\left\|f_{\mathbf{w}+\mathbf{u}}(\mathbf{x})-f_{\mathbf{w}}(\mathbf{x})\right\|_2 & \leq e B \beta^n \sum_i \frac{\left\|\U_i\right\|_2}{\beta} \\
& =e B \beta^{n-1} \sum_i\left\|\U_i\right\|_2 \\
&\leq e^2 n B \tilde{\beta}^{n-1} \sigma \sqrt{2 h \ln ((4+2\sqrt{2}) n h)} \leq \frac{\gamma}{8},
\end{aligned}
\end{equation}
and
\begin{equation}
\label{eq:cod2}
\begin{aligned}
\max _{\mathbf{x} \in \mathcal{X}_B}\left\|f_{\mathbf{w}+\mathbf{u}}(\mathbf{x}+\vv)-f_{\mathbf{w}}(\mathbf{x})\right\|_2 & \leq e (B+\|\vv\|_2) \beta^n \sum_i \frac{\left\|\U_i\right\|_2}{\beta} \\
& =e (B+\|\vv\|_2) \beta^{n-1} \sum_i\left\|\U_i\right\|_2 \\
&\leq e^2 n (B+\sqrt{\tau}\sigma) \tilde{\beta}^{n-1} \sigma \sqrt{2 h \ln ((4+2\sqrt{2}) n h)} \leq \frac{\gamma}{8}.
\end{aligned}
\end{equation}
To make (\ref{eq:cod1}) and (\ref{eq:cod2}) both hold, given $\tilde{\beta}^{n-1} \leq e \beta^{n-1}$, we can choose the largest $\sigma$ (with numerical simplification, i.e., $8e^3\sqrt{2}<2^8$, $4+2\sqrt{2}<8$) as 
\begin{small}
\begin{equation}\nonumber
\sigma^2=\Psi\left(\prod_i\|\W_i\|_2^2\right) = \left(\left(\gamma\Bigg/\left(2^8 n  ( h \ln (8 n h))^\frac{1}{2}\tau^{\frac{1}{2}}\prod_{i=1}^n \|\W_i\|_2^{\frac{n-1}{n}}\right)+\frac{B^2}{4\tau}\right)^{\frac{1}{2}} - \frac{B}{2\tau^{\frac{1}{2}}}\right)^2.
\end{equation}
\end{small}Hence, the perturbation $\mathbf{u}$ with the above value of $\sigma$ satisfies the assumptions of the Lem.~\ref{lem:main1}.
We now compute the KL-term in Lem.~\ref{lem:main1} using the selected distributions for $P$ and $\ul$, considering the given value of $\sigma$.
\begin{equation}\nonumber
\begin{aligned}
\KL(\w+\ul \| P) &\le \frac{\|\w\|_2^2}{2\sigma^2}\\
&=\frac{\sum_{i=1}^n\|\W_i\|_F^2}{2\sigma^2}\\
&\le\mathcal{O}\left( \Phi\left(\prod_i\|\W_i\|_2^2,\sum_i\|\W_i\|_F^2\right) \right),
\end{aligned}
\end{equation}
where
\begin{equation}
\Phi\left(\prod_i\|\W_i\|_2^2,\sum_i\|\W_i\|_F^2\right) = \frac{\sum_{i}\left(\|\W_i\|_F^2 / \|\W_i\|_2^2\right)}{\Psi\left(\prod_i\|\W_i\|_2^2\right) / (\prod_i \|\W\|_2^2)^{\frac{1}{n}}}.
\end{equation}
Clearly, $\Phi$ increases with the growth of $\prod_i\|\W_i\|_2^2$ and $\sum_i\|\W_i\|_F^2$, since
\begin{equation}
\begin{aligned}
&\frac{\partial\Phi}{\partial \left(\prod_i\|\W_i\|_2^2\right)}>0, \\
&\frac{\partial\Phi}{\partial \left(\sum_i\|\W_i\|_F^2\right)}>0,
\end{aligned}
\end{equation}
for all $\prod_i\|\W_i\|_2^2>0$ and $\sum_i\|\W_i\|_F^2>0$.

Then, we can give a union bound over different choices of $\tilde \beta$.
We only need to form the bound for $\left(\frac{\gamma}{2 B}\right)^{\frac{1}{n}} \leq \beta \leq\left(\frac{\gamma \sqrt{m}}{2 B}\right)^{\frac{1}{n}}$ which can be covered using a cover of size $nm^{\frac{1}{2n}}$ as discussed in \citet{neyshabur2017pac}.
Thus, with probability $\ge 1-\delta$, for any $\tilde \beta$ and for all $\w$ such that $|\beta-\tilde{\beta}| \leq \frac{1}{n} \beta$, we have:
\begin{equation}\nonumber
\Lc_{0}(\g) \leq \widehat{\Lc}_{\gamma}(\g)+\mathcal{O}\left ( \sqrt{\frac{\Phi\left(\prod_i\|\W_i\|_2^2,\sum_i\|\W_i\|_F^2\right)+\ln \frac{nm}{\delta}}{m-1}} \right).
\end{equation}

Hence, proved. \hfill $\square$

\end{custompro}

\begin{mylem}[\citet{neyshabur2017pac}]
\label{lem:peturbound}
For any $B, n > 0$, let $f_{\mathbf{w}}: \mathcal{X}_B$ $\rightarrow \mathcal{Y}$ be a $n$-layer feedforward network with ReLU activation function.
Then for any $\w$, and $\x\in \X$, and any perturbation $\ul=\vecc(\{\U_i\}_{i=1}^n)$ such that $\|\U_i\|_2\le \frac{1}{n}\|\W_i\|_2$, the change in the output of the network can be bounded as follow
\begin{equation}
\begin{aligned}
\left\|f_{\mathbf{w}+\mathbf{u}}(\mathbf{x})-f_{\mathbf{w}}(\mathbf{x})\right\|_2 \leq e B\left(\prod_{i=1}^n\left\|\W_i\right\|_2\right) \sum_{i=1}^n \frac{\left\|\U_i\right\|_2}{\left\|\W_i\right\|_2}.
\end{aligned}
\end{equation}
\end{mylem}

\begin{custompro}{for Lem.~\ref{lem:peturbound}.}[\citet{neyshabur2017pac}]
Let $\Delta_i = \|f^i_{\mathbf{w}+\mathbf{u}}(\mathbf{x})-f^i_{\mathbf{w}}(\mathbf{x})\|_2$, we will prove using induction that for any $i\ge 0$:
\begin{equation}
\Delta_i \leq\left(1+\frac{1}{n}\right)^i\left(\prod_{j=1}^i\left\|\W_j\right\|_2\right)\|\mathbf{x}\|_2 \sum_{j=1}^i \frac{\left\|\U_j\right\|_2}{\left\|\W_j\right\|_2}.
\end{equation}
The above inequality together with $(1+\frac{1}{n})^n\le e$ proves the lemma statement. 
The induction base clearly holds since . For any $i\ge 1$, we have the following
\begin{equation}\nonumber
\begin{aligned}
\Delta_{i+1} & =\left\|\left(\W_{i+1}+\U_{i+1}\right) \act\left(f_{\mathbf{w}+\mathbf{u}}^{(i)}(\mathbf{x})\right)-\W_{i+1} \act\left(f_{\mathbf{w}}^{(i)}(\mathbf{x})\right)\right\|_2 \\
& =\left\|\left(\W_{i+1}+\U_{i+1}\right)\left(\act\left(f_{\mathbf{w}+\mathbf{u}}^{(i)}(\mathbf{x})\right)-\act\left(f_{\mathbf{w}}^{(i)}(\mathbf{x})\right)\right)+\U_{i+1} \act\left(f_{\mathbf{w}}^{(i)}(\mathbf{x})\right)\right\|_2 \\
& \leq\left(\left\|\W_{i+1}\right\|_2+\left\|\U_{i+1}\right\|_2\right)\left\|\act\left(f_{\mathbf{w}+\mathbf{u}}^{(i)}(\mathbf{x})\right)-\act\left(f_{\mathbf{w}}^{(i)}(\mathbf{x})\right)\right\|_2+\left\|\U_{i+1}\right\|_2\left\|\act\left(f_{\mathbf{w}}^{(i)}(\mathbf{x})\right)\right\|_2 \\
& \leq\left(\left\|\W_{i+1}\right\|_2+\left\|\U_{i+1}\right\|_2\right)\left\|f_{\mathbf{w}+\mathbf{u}}^{(i)}(\mathbf{x})-f_{\mathbf{w}}^{(i)}(\mathbf{x})\right\|_2+\left\|\U_{i+1}\right\|_2\left\|f_{\mathbf{w}}^{(i)}(\mathbf{x})\right\|_2 \\
& =\Delta_i\left(\left\|\W_{i+1}\right\|_2+\left\|\U_{i+1}\right\|_2\right)+\left\|\U_{i+1}\right\|_2\left\|f_{\mathbf{w}}^{(i)}(\mathbf{x})\right\|_2,
\end{aligned}
\end{equation}
where the last inequality is by the Lipschitz property of the activation function and using $\act(0)=0$.
The $\ell_2$ norm of outputs of layer $i$ is bounded by $\|\mathbf{x}\|_2 \Pi_{j=1}^i\left\|\W_j\right\|_2$ and by the lemma assumption we have $\|\U_{i+1}\|_2\le \frac{1}{n}\|\W_{i+1}\|_2$. 
Therefore, using the induction step, we get the following bound:
\begin{equation}
\begin{aligned}
\Delta_{i+1} & \leq \Delta_i\left(1+\frac{1}{n}\right)\left\|\W_{i+1}\right\|_2+\left\|\U_{i+1}\right\|_2\|\mathbf{x}\|_2 \prod_{j=1}^i\left\|\W_j\right\|_2 \\
& \leq\left(1+\frac{1}{n}\right)^{i+1}\left(\prod_{j=1}^{i+1}\left\|\W_j\right\|_2\right)\|\mathbf{x}\|_2 \sum_{j=1}^i \frac{\left\|\U_j\right\|_2}{\left\|\W_j\right\|_2}+\frac{\left\|\U_{i+1}\right\|_2}{\left\|\W_{i+1}\right\|_2}\|\mathbf{x}\|_2 \prod_{j=1}^{i+1}\left\|\W_i\right\|_2 \\
& \leq\left(1+\frac{1}{n}\right)^{i+1}\left(\prod_{j=1}^{i+1}\left\|\W_j\right\|_2\right)\|\mathbf{x}\|_2 \sum_{j=1}^{i+1} \frac{\left\|\U_j\right\|_2}{\left\|\W_j\right\|_2}.
\end{aligned}
\end{equation}
Hence, proved. \hfill $\square$
\end{custompro}

\begin{mylem}
\label{lam:perbound2}
Given Lem.~\ref{lem:peturbound} and the proof, we have 
\begin{equation}
\left\|f_{\mathbf{w}+\mathbf{u}}(\mathbf{x}+\vv)-f_{\mathbf{w}}(\mathbf{x})\right\|_2 \leq e (B+\|\vv\|_2)\left(\prod_{i=1}^n\left\|\W_i\right\|_2\right) \sum_{i=1}^n \frac{\left\|\U_i\right\|_2}{\left\|\W_i\right\|_2}.
\end{equation}
\end{mylem}

\subsection{Proof for \Cref{thm:main3}}
\label{app:thm3.3}

First, we restate \Cref{thm:main3} here.
\begin{customthm}{3.3}
Given \Cref{thm:main2}, 
for any $\x\in \X$, suppose there exist $p^A_{\w}(\x)$, $p^B_{\w}(\x)$ such that
\begin{equation}\nonumber
\begin{aligned}
    &\E_{\ul} \mathbbm{1} \Big[\underset{c}{\argmax} f_{\w+\ul}(\x+\vv)[c]=\g_{0,\w}(\x)\Big] \ge p^A_{\w}(\x) \ge p^B_{\w}(\x)\\
    &\quad\quad\quad\quad\quad\quad\quad\quad\quad\quad\quad\quad\quad\quad\quad\quad\quad\quad\ge 
    \max_{j\ne \g_{0,\w}(\x)} \E_{\ul} \mathbbm{1} \Big[ \underset{c}{\argmax} f_{\w+\ul}(\x+\vv)[c]=j \Big].
\end{aligned}
\end{equation}
Then, for any $\delta,\gamma>0$, with probability at least $1-\delta$ we have
\begin{equation}\nonumber
\Lc_0(\g,\epsilon) 
\leq \widehat{\Lc}_{\gamma}(\g)+\mathcal{O}\left ( \sqrt{\frac{\Phi\left(\prod_i\|\W_i\|_2^2,\sum_i\|\W_i\|_F^2\right)+\ln \frac{nm}{\delta}}{m-1}} \right)
\end{equation}
within $\ell_2$ norm data perturbation radius $\sqrt{\epsilon_{\x}}$, where
\begin{equation}\nonumber
\epsilon_{\x} = 
\underbrace{-\ln\left(1-\left(\sqrt{p^A_{\w}(\x)}-\sqrt{p^B_{\w}(\x)}\right)^2\right)}_{\textbf{Model and Data Joint Dependence}} 
\;\cdot\;
2 \underbrace{\Psi\left(\prod_i\|\W_i\|_2^2\right)}_{\textbf{Model Dependence}}.
\end{equation}
\end{customthm}

In the following, we provide the details of the proof.
This proof is developed from \citet{dvijotham2020framework,xie2021crfl}, but  their smoothing function is only applied on $\x$ or $\w$, our certification is with respect to the smoothed Majority Vote classifier      
               .

\begin{custompro}{for \Cref{thm:main3}}
Consider a classifier $f_\w:\X\to\mathcal{Y}$. 
The output of the classifier relies on both the input $\x$ and its model weights $\w$.
We would like to verify the robustness of the smoothed Majority Vote classifier $\g$.
Recall that the smoothed Majority Vote classify is randomized on both $\x$ with $\vv$ and $\w$ with $\ul$, to prove \Cref{thm:main3}, our goal is to certify that
\begin{equation}
\label{eq:certify}
    \g_{0,\w}(\x) = \g_{0,\w}(\x+\varp)
\end{equation}
for all $\varp\in\{\varp\in\R^d \;|\; \|\varp\|_2^2\le \epsilon_{\x} \}$ and  all $\x\in \X$, 
where $\epsilon_{\x}$ satisfies the condition in \Cref{thm:main3}.

We let $D_g$ be g-divergence (as we have used $f(\cdot)$, we define g-divergence rather than f-divergence), $\epsilon_g(\x)=D_g(\nu\| \rho)$ where $\nu$ is the joint distribution of $\w+\ul$ and $\x+\varp+\vv$ with probability density function (PDF) $\nu(\cdot,\cdot)$, $\rho$ is the joint distribution of $\w+\ul$ and $\x+\vv$ with PDF $\rho(\cdot,\cdot)$, $p_{\w,\ul}^A(\x)$ and $p^B_{\w}(\x)$ are as in \Cref{thm:main3}. 
Let
\begin{equation}
\label{eq:PaPb}
\begin{aligned}
&\mathcal{P}_{\w}^A(\x)=\E_{\ul,\vv} \mathbbm{1} \Big[ \underset{c}{\argmax} f_{\w+\ul}(\x+\vv)[c]=\g_{0,\w}(\x)\Big], \\
&\mathcal{P}_{\w}^B(\x)=\max_{j\ne \g_{0,\w}(\x)} \E_{\ul,\vv} \mathbbm{1} \Big[ \underset{c}{\argmax} f_{\w+\ul}(\x+\vv)[c]=j \Big],\\
&\mathcal{P}_{\w}^A(\x)+\mathcal{P}_{\w}^B(\x)\le 1,
\end{aligned}
\end{equation}
and
\begin{equation}
\label{eq:ineq111}
    \mathcal{P}_{\w}^A(\x) \ge p^A_{\w}(\x)\ge p^B_{\w}(\x) \ge 
    \mathcal{P}_{\w}^B(\x).
\end{equation}
According to Pf.~\ref{app:pfa6}, we have that: the smoothed classifier $\g_{0,\w}(\x)$ is robustly certified, i.e., (\ref{eq:certify}) holds,
if the optimal value of the following convex optimization problem is non-negative, i.e.,
\begin{equation}
\label{eq:optimization11}
\max _{\lambda \geq 0, \kappa} \kappa-\lambda \epsilon_g(\x)-\mathcal{P}_{\w}^A(\x) g_\lambda^*(\kappa-1)-\mathcal{P}^B_{\w}(\x) g_\lambda^*(\kappa+1) - (1-\mathcal{P}_{\w}^A(\x)-\mathcal{P}_{\w}^B(\x)) g_\lambda^*(\kappa) \geq 0,
\end{equation}
where $g_\lambda^*(u)=\max _{v \geq 0}(u v-g_\lambda(v)), g_\lambda(v)=\lambda g(v)$, the function $g(\cdot)$ is used in g-divergence. 

Then, let $D_g$ be $\KL$ divergence, according to Pf.~\ref{app:pfa7}, the optimization problem of (\ref{eq:optimization11}) is non-negative if 
\begin{equation}
\label{eq:opt1500}
\KL(\nu\|\rho) \leq-\ln \left(1-\left(\sqrt{\mathcal{P}^A_{\w}(\x)}-\sqrt{\mathcal{P}^B_{\w}(\x)}\right)^2\right).
\end{equation}
Since $\ul, \vv\sim \mathcal{N}(0,\sigma^2 \mathbf{I})$, we have
\begin{equation}
\label{eq:kldetail}
\begin{aligned}
\KL(\nu\|\rho) &= \KL\left( 
\left(\begin{smallmatrix}\w+\ul \\\x+\varp+\vv \end{smallmatrix}\right) \Big\|
\left(\begin{smallmatrix}\w+\ul \\\x+\vv \end{smallmatrix}\right)
\right)\\
&=\frac{\|\left(\begin{smallmatrix}\w \\\x+\varp \end{smallmatrix}\right)-\left(\begin{smallmatrix}\w \\\x\end{smallmatrix}\right)\|_2^2}{2\sigma^2} \\
&=\frac{\|\varp\|_2^2}{2\sigma^2}.
\end{aligned}
\end{equation}
Given \Cref{thm:main2}, (\ref{eq:ineq111}), (\ref{eq:opt1500}) and (\ref{eq:kldetail}), we have that $\g_{0,\w}(\x)$ is certified robust if 
\begin{equation}
\|\varp\|_2^2 \le \epsilon_{\x} = 
\underbrace{-\ln\left(1-\left(\sqrt{p^A_{\w}(\x)}-\sqrt{p^B_{\w}(\x)}\right)^2\right)}_{\textbf{Data Dependence}} 
\;\cdot\;
2 \underbrace{\Psi\left(\prod_i\|\W_i\|_2^2\right)}_{\textbf{Model Dependence}}.
\end{equation}
Hence, proved. \hfill $\square$
\end{custompro}

\begin{mypro}
\label{app:pfa6}
Let $D_g$ be g-divergence, 
the function $g(\cdot)$ is used in g-divergence,
$\epsilon_g(\x)=D_g(\nu\| \rho)$ where $\nu$ is the joint distribution of $\w+\ul$ and $\x+\varp+\vv$ with PDF $\nu(\ww,\xx)$, $\rho$ is the joint distribution of $\w+\ul$ and $\x+\vv$ with PDF $\rho(\ww,\xx)$, $p_{\w,\ul}^A(\x)$ and $p^B_{\w}(\x)$ be as in \Cref{thm:main3}. 
Let $r(\ww,\xx)=\frac{\nu(\ww,\xx)}{\rho(\ww,\xx)}$ be likelihood ratio, and  
\begin{equation}
\phi(\ww,\xx)= \begin{cases}+1, & \text { if } 
\argmax\limits_c 
f_{\ww} (\xx)[c] =\g_{0,\w}(\x) \\
-1, & \text { else if } \argmax\limits_c f_{\ww}(\xx)[c] =\underset{j\ne \g_{0,\w}(\x)}{\max} \E_{\ul,\vv} \mathbbm{1} \Big[ \argmax\limits_c f_{\w+\ul}(\x+\vv)[c]=j \Big]\\
0, & \text{ otherwise}
\end{cases}
\end{equation}
we have
\begin{equation}
\begin{aligned}
\mathbb{E}_{(\ww, \xx) \sim \nu}[\phi(\ww,\xx)] & =\mathbb{E}_{(\ww, \xx) \sim \rho}[r(\ww,\xx) \phi(\ww,\xx)], \\
D_g(\nu \| \rho) & =\mathbb{E}_{(\ww,\xx) \sim \rho}[g(r(\ww,\xx))], \\
\mathbb{E}_{(\ww,\xx) \sim \rho}[r(\ww,\xx)] & =1.
\end{aligned}
\end{equation}
The third condition is obtained using the fact that $\nu$ is a probability measure.
The optimization over $\nu$, which is equivalent to optimize over $r$, can be rewritten as
\begin{equation}
\label{eq:optimize333}
\begin{aligned}
& \min_{r \geq 0} \mathbb{E}_{(\ww,\xx)\sim\rho}[r(\ww,\xx) \phi(\ww,\xx)] \\
& \; s.t. \; \mathbb{E}_{(\ww,\xx) \sim \rho}[g(r(\ww,\xx))] \leq \epsilon_g(\x), \;\mathbb{E}_{(\ww,\xx)\sim \rho}[r(\ww,\xx)]=1.
\end{aligned}
\end{equation}
We solve the optimization using Lagrangian duality as follows.
We first dualize the constraints on $r$ to obtain
\begin{equation}
\begin{aligned}
& \min _{r \geq 0} \mathbb{E}_{(\ww,\xx) \sim \rho}[r(\ww,\xx) \phi(\ww,\xx)]+\lambda\left(\mathbb{E}_{(\ww,\xx) \sim \rho}[g(r(\ww,\xx))]-\epsilon_g(\x)\right)+\kappa\left(1-\mathbb{E}_{(\ww,\xx) \sim \rho}[r(\ww,\xx)]\right) \\
& =\min _{r \geq 0} \mathbb{E}_{(\ww,\xx) \sim \rho}[r(\ww,\xx) \phi(\ww,\xx)+\lambda g(r(\ww,\xx))-\kappa r(\ww,\xx)]+\kappa-\lambda \epsilon_g(\x). 
\end{aligned}
\end{equation}
As the $\ww$ components of $\nu$ and $\rho$ are identical, but $\xx$ components of $\nu$ and $\rho$ can be different, let $r(\xx)=\E_{\ww\sim\rho_{\ww}}[r(\ww,\xx)]$ where $\rho_{\ww}$ is the marginal distribution of $\ww$ for $\rho$,
the above optimization problem can be rewritten as
\begin{equation}
\label{eq:optimize222}
\begin{aligned}
&\kappa-\lambda \epsilon_g(\x)-\mathbb{E}_{\xx \sim \rho_{\xx}}\left[\max _{r \geq 0} \kappa r(\xx)-r(\xx) \mathbb{E}_{\ww\sim\rho_{\ww}}\phi(\ww,\xx)-\lambda g(r(\xx))\right] \\
& =\kappa-\lambda \epsilon_g(\x)-\mathbb{E}_{\xx \sim \rho_{\xx}}\left[\max _{r \geq 0} r(\xx)(\kappa-\mathbb{E}_{\ww\sim\rho_{\ww}}\phi(\ww,\xx))-\lambda g(r(\xx))\right] \\
& =\kappa-\lambda \epsilon_g(\x)-\mathbb{E}_{\xx \sim \rho_{\xx}}\left[\max _{r \geq 0} r(\xx)(\kappa-\mathbb{E}_{\ww\sim\rho_{\ww}}\phi(\ww,\xx))-g_\lambda(r(\xx))\right] \\
& = \kappa-\lambda \epsilon_g(\x)-\mathbb{E}_{\xx \sim \rho_{\xx}}\left[g_\lambda^*(\kappa-\mathbb{E}_{\ww\sim\rho_{\ww}}\phi(\ww,\xx))\right],
\end{aligned}
\end{equation}
where $g_\lambda^*(u)=\max _{v \geq 0}(u v-g_\lambda(v)), g_\lambda(v)=\lambda g(v)$. 
Since strong duality, we can maximize the final term in (\ref{eq:optimize222}) with respect to $\lambda\ge 0$, $\kappa$, to achieve the optimal value in (\ref{eq:optimize333}). 
If the optimal value, i.e.,
\begin{equation}
\label{eq:opt55555}
\max_{\lambda\ge0,\kappa} \kappa-\lambda \epsilon_g(\x)-\mathbb{E}_{\xx \sim \rho_{\xx}}\left[g_\lambda^*(\kappa-\mathbb{E}_{\ww\sim\rho_{\ww}}\phi(\ww;\xx))\right]
\end{equation}
is non-negative, then (\ref{eq:certify}) holds.
As we have
\begin{equation}
\label{eq:opt666666}
\begin{aligned}
&\max_{\lambda\ge0,\kappa} \kappa-\lambda \epsilon_g(\x)-\mathbb{E}_{\xx \sim \rho_{\xx}}\left[g_\lambda^*(\kappa-\mathbb{E}_{\ww\sim\rho_{\ww}}\phi(\ww;\xx))\right]\\
&\ge \max_{\lambda\ge0,\kappa} \kappa-\lambda \epsilon_g(\x)-\mathbb{E}_{(\ww,\xx) \sim \rho}\left[g_\lambda^*(\kappa-\phi(\ww;\xx))\right],
\end{aligned}
\end{equation}
given (\ref{eq:PaPb}), (\ref{eq:opt55555}), (\ref{eq:opt666666}), we get that (\ref{eq:certify}) holds if 
\begin{equation}
\max _{\lambda \geq 0, \kappa} \kappa-\lambda \epsilon_g(\x)-\mathcal{P}_{\w}^A(\x) g_\lambda^*(\kappa-1)-\mathcal{P}^B_{\w}(\x) g_\lambda^*(\kappa+1) - (1-\mathcal{P}_{\w}^A(\x)-\mathcal{P}_{\w}^B(\x)) g_\lambda^*(\kappa) \geq 0.
\end{equation}
\hfill $\square$
\end{mypro}

\begin{mypro}
\label{app:pfa7}
We use the KL divergence function $g(u) = u\ln(u)$ for $D_g$, which is a convex function with $g(1) = 0$.
Thus, we have
\begin{equation}
g_\lambda^*(u)=\max _{v \geq 0}(u v-\lambda g(v))=\max _{v \geq 0}(u v-\lambda v \log (v)).
\end{equation}
Use the derivative with respect to $v$ to 0 to solve the above optimization problem, i.e.,
\begin{equation}
\frac{\partial(uv-\lambda v\ln(v))}{\partial v}=0,
\end{equation}
we have $v=\ln\frac{u-\lambda}{\lambda}$, $\lambda>0$.
Thus we get
$$
g_\lambda^*(u)=\lambda \exp \left(\frac{u}{\lambda}-1\right).
$$
Suppose there exists a bound $\epsilon_{\mathrm{KL}}(\x)$ on the $\mathrm{KL}$ divergence, i.e., $\KL(\nu \| \rho) \leq \epsilon_{\mathrm{KL}}(\x)$, then the optimization problem in (\ref{eq:optimization11}) can be rewritten as
\begin{equation}
\label{eq:opt8888}
\begin{aligned}
\max _{\lambda>0, \kappa}\Bigg(\kappa&-\lambda \epsilon_{\mathrm{KL}}(\x)-\mathcal{P}^A_{\w}(\x) \lambda \exp \left(\frac{\kappa-1}{\lambda}-1\right)-\mathcal{P}^B_{\w}(\x) \lambda \exp \left(\frac{\kappa+1}{\lambda}-1\right)\\
&\quad\quad\quad\quad\quad\quad\quad\quad\quad\quad\quad\quad\quad-(1-\mathcal{P}^A_{\w}(\x)-\mathcal{P}^B_{\w}(\x)) \lambda \exp \left(\frac{\kappa}{\lambda}-1\right)
\Bigg) \geq 0 .
\end{aligned}
\end{equation}
Let $\xi=\kappa / \lambda, \zeta=\frac{1}{\lambda}$ (with $\zeta>0$), ($\ref{eq:opt8888}$) can be rewritten as:
\begin{equation}
\begin{aligned}
\max _{\zeta>0, \xi}\Bigg(\frac{1}{\zeta}\Big(\xi-\epsilon_{\mathrm{KL}}(\x)&-\mathcal{P}^A_{\w}(\x) \exp (\xi-\zeta-1)-\mathcal{P}^B_{\w}(\x) \exp (\xi+\zeta-1)\\
&\quad\quad\quad\quad\quad\quad\quad\quad\quad-\left(1-\mathcal{P}^A_{\w}(\x)-\mathcal{P}^B_{\w}(\x)\right) \exp (\xi-1)\Big)\Bigg) \geq 0.
\end{aligned}
\end{equation}
Since $\zeta>0$, let both left-hand side and right-hand side time $\zeta$ and the above optimization problem can be rewritten as:
\begin{equation}
\label{eq:opt9999}
\begin{aligned}
\max _{\zeta>0, \xi}\Bigg(\xi-\epsilon_{\mathrm{KL}}(\x)&-\mathcal{P}^A_{\w}(\x) \exp (\xi-\zeta-1)-\mathcal{P}^B_{\w}(\x) \exp (\xi+\zeta-1)\\
&\quad\quad\quad\quad\quad\quad\quad\quad\quad\quad-\left(1-\mathcal{P}^A_{\w}(\x)-\mathcal{P}^B_{\w}(\x)\right) \exp (\xi-1)\Bigg) \geq 0.
\end{aligned}
\end{equation}
Setting the derivative of the left-hand side with respect to $\zeta$ to $0$ and solving for $\zeta$, we obtain
\begin{equation}
\label{eq:opt1000}
\begin{aligned}
&\mathcal{P}^A_{\w}(\x) \exp (\xi-\zeta-1)-\mathcal{P}^B_{\w}(\x) \exp (\xi+\zeta-1)  =0, \\
&\zeta =\ln \left(\sqrt{\frac{\mathcal{P}^A_{\w}(\x)}{\mathcal{P}^B_{\w}(\x)}}\right) .
\end{aligned}
\end{equation}
Given (\ref{eq:opt9999}) and (\ref{eq:opt1000}), we have
\begin{equation}
\label{eq:opt1100}
\max_\xi \left(\xi-\epsilon_{\mathrm{KL}}(\x)-\left(1-\left(\sqrt{\mathcal{P}^A_{\w}(\x)}-\sqrt{\mathcal{P}^B_{\w}(\x)}\right)^2\right) \exp (\xi-1)\right) \ge 0.
\end{equation}
Setting the derivative with respect to $\xi$ to $0$ and solving for $\xi$, we obtain
\begin{equation}
\label{eq:opt1200}
\begin{aligned}
&1-\left(1-\left(\sqrt{\mathcal{P}^A_{\w}(\x)}-\sqrt{\mathcal{P}^B_{\w}(\x)}\right)^2\right) \exp (\xi-1) =0, \\
&\xi =1-\ln \left(1-\left(\sqrt{\mathcal{P}^A_{\w}(\x)}-\sqrt{\mathcal{P}^B_{\w}(\x)}\right)^2\right) .
\end{aligned}
\end{equation}
Given (\ref{eq:opt1100}) and (\ref{eq:opt1200}), we have
\begin{equation}
\begin{aligned}
&-\ln \left(1-\left(\sqrt{\mathcal{P}^A_{\w}(\x)}-\sqrt{\mathcal{P}^B_{\w}(\x)}\right)^2\right)-\epsilon_{\mathrm{KL}}(\x) \ge 0 .
\end{aligned}
\end{equation}
Then, we have
\begin{equation}
\KL(\nu\|\rho) \leq-\ln \left(1-\left(\sqrt{\mathcal{P}^A_{\w}(\x)}-\sqrt{\mathcal{P}^B_{\w}(\x)}\right)^2\right).
\end{equation}
Hence, proved. \hfill $\square$
\end{mypro}

\hspace*{\fill}

\hspace*{\fill}

\subsection{Prop.~\ref{prop:5.2}}
\label{app:prop5.2}

In this work, we adopt the following Prop.~\ref{prop:5.2} to evaluate the certified radius for smoothed Majority Vote classifiers.
\begin{customprop}{5.2}
\label{prop:5.2}
Let $f_{\mathbf{w}}: \mathcal{X}_B$ $\rightarrow \mathcal{Y}$ be the base classifier of $\g$, the perturbations $\ul,\vv \sim \mathcal{N}(0,\sigma^2 \mathbf{I})$.
For any $\x\in \X$, suppose there exists $p^A_{\w}(\x)$, $p^B_{\w}(\x)$ such that
\begin{equation}\nonumber
\begin{aligned}
    &\E_{\ul} \mathbbm{1} \Big[\underset{c}{\argmax} f_{\w+\ul}(\x+\vv)[c]=\g_{0,\w}(\x)\Big] \ge p^A_{\w}(\x) \ge p^B_{\w}(\x)\\
    &\ge 
    \max_{j\ne \g_{0,\w}(\x)} \E_{\ul} \mathbbm{1} \Big[ \underset{c}{\argmax} f_{\w+\ul}(\x+\vv)[c]=j \Big].
\end{aligned}
\end{equation}
Then, we have $\g_{0,\w}(\x+\varp)=\g_{0,\w}(\x)$ for all $\|\varp\|_2\le R$, where
\begin{equation}\nonumber
R^2 = 
-2\sigma^2 \ln\left(1-\left(\sqrt{p^A_{\w}(\x)}-\sqrt{p^B_{\w}(\x)}\right)^2\right).
\end{equation}
\end{customprop}
In the following, we provide the details of the proof.

\begin{custompro}{for Prop.~\ref{prop:5.2}}
We would like to verify the robustness of the smoothed Majority Vote classifier $\g$.
Recall that $\g$ is randomized on both $\w$ and $\x$ with $\ul, \vv\sim \mathcal{N}(0,\sigma^2 \mathbf{I})$, to prove Prop.~\ref{prop:5.2}, our goal is to certify that
\begin{equation}
\label{eq:certify2}
    \g_{0,\w}(\x) = \g_{0,\w}(\x+\varp)
\end{equation}
for all $\varp\in\{\varp\in\R^d \;|\; \|\varp\|_2\le R \}$, 
where
\begin{equation}\nonumber
R^2 = 
-2\sigma^2 \ln\left(1-\left(\sqrt{p^A_{\w}(\x)}-\sqrt{p^B_{\w}(\x)}\right)^2\right).
\end{equation}

We let $D_g$ be g-divergence (as we have used $f(\cdot)$, we define g-divergence rather than f-divergence), $\epsilon_g(\x)=D_g(\nu\| \rho)$ where $\nu$ is the joint distribution of $\w+\ul$ and $\x+\varp+\vv$ with PDF $\nu(\cdot,\cdot)$, $\rho$ is the joint distribution of $\w+\ul$ and $\x+\vv$ with PDF $\rho(\cdot,\cdot)$, $p_{\w,\ul}^A(\x)$ and $p^B_{\w}(\x)$ be as in Prop.~\ref{prop:5.2}. 
Let
\begin{equation}
\label{eq:PaPb2}
\begin{aligned}
&\mathcal{P}_{\w}^A(\x)=\E_{\ul,\vv} \mathbbm{1} \Big[ \underset{c}{\argmax} f_{\w+\ul}(\x+\vv)[c]=\g_{0,\w}(\x)\Big], \\
&\mathcal{P}_{\w}^B(\x)=\max_{j\ne \g_{0,\w}(\x)} \E_{\ul,\vv} \mathbbm{1} \Big[ \underset{c}{\argmax} f_{\w+\ul}(\x+\vv)[c]=j \Big],\\
&\mathcal{P}_{\w}^A(\x)+\mathcal{P}_{\w}^B(\x)\le 1,
\end{aligned}
\end{equation}
and
\begin{equation}
\label{eq:ineq111222}
    \mathcal{P}_{\w}^A(\x) \ge p^A_{\w}(\x)\ge p^B_{\w}(\x) \ge 
    \mathcal{P}_{\w}^B(\x).
\end{equation}
According to Pf.~\ref{app:pfa6}, we have that: the smoothed classifier $\g_{0,\w}(\x)$ is robustly certified, i.e., (\ref{eq:certify2}) holds,
if the optimal value of the following convex optimization problem is non-negative, i.e.,
\begin{equation}
\label{eq:optimization1122}
\max _{\lambda \geq 0, \kappa} \kappa-\lambda \epsilon_g(\x)-\mathcal{P}_{\w}^A(\x) g_\lambda^*(\kappa-1)-\mathcal{P}^B_{\w}(\x) g_\lambda^*(\kappa+1) - (1-\mathcal{P}_{\w}^A(\x)-\mathcal{P}_{\w}^B(\x)) g_\lambda^*(\kappa) \geq 0,
\end{equation}
where $g_\lambda^*(u)=\max _{v \geq 0}(u v-g_\lambda(v)), g_\lambda(v)=\lambda g(v)$, the function $g(\cdot)$ is used in g-divergence. 

Then, let $D_g$ be $\KL$, according to Pf.~\ref{app:pfa7}, the optimization problem of (\ref{eq:optimization1122}) is non-negative if 
\begin{equation}
\label{eq:opt150022}
\KL(\nu\|\rho) \leq-\ln \left(1-\left(\sqrt{\mathcal{P}^A_{\w}(\x)}-\sqrt{\mathcal{P}^B_{\w}(\x)}\right)^2\right).
\end{equation}
Since $\ul,\vv\sim \mathcal{N}(0,\sigma^2 \mathbf{I})$, we have
\begin{equation}
\label{eq:kldetail2}
\begin{aligned}
\KL(\nu\|\rho) &= \KL\left( 
\left(\begin{smallmatrix}\w+\ul \\\x+\varp+\vv \end{smallmatrix}\right) \Big\|
\left(\begin{smallmatrix}\w+\ul \\\x+\vv \end{smallmatrix}\right)
\right)\\
&=\frac{\|\left(\begin{smallmatrix}\w \\\x+\varp \end{smallmatrix}\right)-\left(\begin{smallmatrix}\w \\\x\end{smallmatrix}\right)\|_2^2}{2\sigma^2} \\
&=\frac{\|\varp\|_2^2}{2\sigma^2}.
\end{aligned}
\end{equation}
Given (\ref{eq:ineq111222}), (\ref{eq:opt150022}) and (\ref{eq:kldetail2}), we have that $\g_{0,\w}(\x)$ is certified robust if 
\begin{equation}
\|\varp\|_2 \le R = \sqrt{ 
-2\sigma^2\ln\left(1-\left(\sqrt{p^A_{\w}(\x)}-\sqrt{p^B_{\w}(\x)}\right)^2\right)}.
\end{equation}
Hence, proved. \hfill $\square$
\end{custompro}

\section{Experimental details}
\label{app:experimental details}

\subsection{Details of models}
\label{app:model}
We follow the training scheme at \citet{salman2019provably}, 
the details of the model are given in the following.

For MNIST and Fashion-MNIST, we adopt a $(28\times 28)\times 32 \times 32 \times 32 \times 10$ fully connected neural network with ReLU activation function. The models are trained for 30 epochs using SGD with momentum 0.9, batch size 256, smoothing noise variance 0.12, and an initial learning rate of 0.1 that is divided by 10 at the 10th and 20th epochs, with/without our regularizer ($\alpha=0.1$).
For CIFAR10, we adopt a ResNet110 with ReLU activation function. The model is trained for 150 epochs using SGD with momentum 0.9, batch size 256, smoothing noise variance 0.12, and an initial learning rate of 0.1 that is divided by 10 at the 50th and 100th epochs, with/without our regularizer ($\alpha=0.1$).

For ImageNet, we adopt a ResNet50 with ReLU activation function. We finetune the pretrained model in \url{https://github.com/Hadisalman/smoothing-adversarial} for 10 epochs using SGD with momentum 0.9, batch size 128, smoothing noise variance 0.25, and a learning rate of 0.0001, with/without our regularizer ($\alpha=0.1$).

We conduct all experiments on a server with 4 RTX 4090 GPUs.

\subsection{Sharpness-like method}
\label{app:sharpness}

Sharpness of the local minima captures the sensitivity/stability of the empirical risk to perturbations in model weights. Recently, it becomes a popular generalization measure \citep{jiang2019fantastic,dziugaite2020in}, and is also captured elegantly by some PAC-Bayesian frameworks which add randomly generated perturbations to the weights as posterior.

In this study, extending the approach in \citet{neyshabur2017pac}, we also establish a connection between sharpness and weight spectral norm. Within the output random perturbation limitation (also referred to as sharpness limitation, as detailed in \Cref{lem:main1}), we find that a lower weight spectral norm in $\Psi(\prod_i\|{\bf W}_i\|_2^2)$ can result in a higher $\sigma^2$. 
This, in turn, leads to both a tighter robust generalization bound and a larger certified radius, as provided in \Cref{thm:main3}. 
However, experimentally, directly utilizing $\Psi(\prod_i\|{\bf W}_i\|_2^2)$ to estimate $\sigma^2$ is difficult due to other parameters in the formula, such as the margin $\gamma$. 
We note that a large random perturbation in model output normally indicates a high likelihood of reversed classification. Thus, the theoretically largest $\sigma^2$ as suggested by $\Psi(\prod_i\|{\bf W}_i\|_2^2)$ under an output perturbation limitation can be practically approximated by the largest $\sigma^2$ under a deviation in training accuracy, as proposed in \citet{jiang2019fantastic}. Therefore, following \citet{jiang2019fantastic}, we employ a sharpness-like method to estimate $\sigma^2$ as follows.

\noindent
\textbf{Method:} Collect 50 samples, ${\bf u}_ 1$, ..., ${\bf u}_ {50}$, from a Gaussian distribution $\mathcal{N}(0,\sigma^2 \bf I)$. Then find the largest $\sigma^2$ (from 0.01, 0.02, to 1) that ensures the training accuracy drop between the base classifier $f_{\bf w}$ and the perturbed classifiers $f_ {{\bf w}+{\bf u_i}}$ remains within an acceptable tolerance limit. For the MNIST dataset, this tolerance is 2\%, while for other datasets, it is 5\%.

\begin{myrem}
Theoretically $\Psi(\prod_i\|{\bf W}_i\|_2^2)$-approximated $\sigma^2$ and empirically sharpness-method-estimated $\sigma^2$ can both capture the sharpness (flatness) of the model, and they are highly correlated in our experiments (small weight spectral norm (\Cref{fig:2}) leads to large sharpness-method-estimated $\sigma^2$ and large certified radius (\Cref{fig:3})) and the experiments in \citet{jiang2019fantastic}. 
Therefore, for experimental convenience, we employ a widely-used sharpness-like method to estimate $\sigma^2$.
\end{myrem}

\subsection{Details of certified radius in the experiments}
\label{app:radius}
Firstly, we define the certified robust test error (accuracy).
Let $\mathcal{S}_ {test}$ be the test set, $m_ {test}$ be the size of test set, and $r_{test}$ be the $\ell_2$ norm perturbation radius, the certified test error under $\ell_2$ norm perturbation radius $r_{test}$ is defined as $$\frac{1}{m_ {test}}\underset{({\bf x},y) \in \mathcal{S}_ {test}}{\sum} \mathbbm{1} \Big[ \exists \varp \Big| \|\varp\|_2\le r_{test}, \mathcal{G} _ {0, {\bf w}}({\bf x}+\varp) \ne y \Big].$$

It is difficult to directly compute the above certified test error.
Thus, in this work, our primary evaluation algorithm closely adheres to \citet{cohen2019certified}, to estimate certified radius $R$ for each test sample, it allows us to evaluate certified test error (accuracy) under any radius.
There exists two minor differences between ours and \citet{cohen2019certified}.

The first difference lies in the certified robust radius, our randomized smoothing evaluation framework  (see \Cref{app:prop5.2}) computes certified radius for each sample through 
\begin{equation}
R = \sqrt{ 
-2\sigma^2\ln\left(1-\left(\sqrt{p^A_{\w}(\x)}-\sqrt{p^B_{\w}(\x)}\right)^2\right)},
\end{equation}
while \citet{cohen2019certified} uses the inverse cumulative distribution function of Gaussian.

The second difference lies in the functions to estimate $p_ {\bf w}^A({\bf x})$ and $p_ {\bf w}^B({\bf x})$. 
Following \citet{cohen2019certified}, we adapt the function SAMPLEUNDERNOISE$(f_ {\bf w}, {\bf x}, num, \sigma)$, then utilize the CERTIFY function to estimate the lower bound of $p_ {\bf w}^A({\bf x})$ and approximate the upper bound of $p_ {\bf w}^B({\bf x})$ through $1-p_ {\bf w}^A({\bf x})$. 
Please note that, mirroring the approach in \citet{cohen2019certified}, we employ the lower bound of $p_ {\bf w}^A({\bf x})$ and the upper bound of $p_ {\bf w}^B({\bf x})$ to estimate the certified radius (which can also be regarded as a lower bound of the certified radius). 
In the following, we introduce three sequential functions,  SAMPLEUNDERNOISE,  LOWERCONFBOUND, and CERTIFY from \citet{cohen2019certified}.

\noindent
\textbf{1. Updated  SAMPLEUNDERNOISE:}

We redesign the function for the smoothed Majority Vote classifier $\g$.
\begin{enumerate}
    \item Draw $num$ samples of noise, $\ul_1,...,\ul_{num}\sim\mathcal{N}(0,\sigma^2\mathbf{I})$, $\vv_1,...,\vv_{num}\sim\mathcal{N}(0,\sigma^2\mathbf{I})$.
    \item Run the noisy images and noisy weights through the base classifier $f_\w$ to obtain the predictions $f_{\w+\ul_1}(\x+\vv_1),...,f_{\w+\ul_{num}}(\x+\vv_{num})$.
    \item Return the counts for each class.
\end{enumerate}
Here, we sample the smoothed Majority Vote classifier with $\x+\vv_i$ and $\w+\ul_i$, and return the majority class.  

\noindent
\textbf{2. LOWERCONFBOUND:}

Following \citet{cohen2019certified}, the function LOWERCONFBOUND$(k, n, 1-\alpha_b)$ returns a one-sided $(1-\alpha_b)$ lower confidence interval for the Binomial parameter $p$ given a sample $k\sim$Binomial$(n, p)$.

\noindent
\textbf{3. CERTIFY:}
\begin{enumerate}
    \item $counts0$ $\gets$ SAMPLEUNDERNOISE$(f_ {\bf w}, {\bf x}, num_0, \sigma)$
    \item $\hat c^A$ $\gets$ top index in $counts0$
    \item $counts$ $\gets$ SAMPLEUNDERNOISE$(f_ {\bf w}, {\bf x}, num, \sigma)$
    \item (lower bound of $p_ {\bf w}^A$)$\gets$ LOWERCONFBOUND$(counts[\hat c^A], num, 1-\alpha_b)$
    \item if (lower bound of $p_ {\bf w}^A$)$>0.5$, return it; else return ABSTAIN.
\end{enumerate}

When running the CERTIFY algorithm, we use 100 ($num_0$) Monte Carlo samples for selection, 100000 ($num$) samples for estimation on MNIST and FashionMNIST, 10000 samples ($num$) for estimation on CIFAR10 and ImageNet, and $\alpha_b=0.001$.

\vskip 0.2in
\bibliography{sample}

\end{document}